\documentclass{article}

\PassOptionsToPackage{dvipsnames}{xcolor}
\PassOptionsToPackage{hyperfootnotes=false}{hyperref}
\usepackage{xcolor}
\usepackage[final]{corl_2026} 
\hypersetup{
  pdftitle={Bimanual Robot Manipulation via Multi-Agent In-Context Learning},
  pdfauthor={Alessio Palma, Indro Spinelli, Vignesh Prasad, Luca Scofano, Yufeng Jin, Georgia Chalvatzaki, Fabio Galasso}
}

\title{Bimanual Robot Manipulation via Multi-Agent In-Context Learning}

\author{
  \textbf{Alessio Palma$^{1}$\thanks{\texttt{alessio.palma@uniroma1.it}},
  Indro Spinelli$^{1}$,
  Vignesh Prasad$^{2,3,4}$,
  Luca Scofano$^{1}$,
  Yufeng Jin$^{2}$,}\\
  \textbf{Georgia Chalvatzaki$^{2,3,4}$\thanks{Co-senior authors.},
  and Fabio Galasso$^{1}$\footnotemark[2]}\\
  $^{1}$PINlab, Sapienza University of Rome, Italy\\
  $^{2}$Interactive Robot Perception \& Learning (PEARL) Lab, TU Darmstadt, Germany\\
  $^{3}$Hessian.AI, Germany\hspace{2em}
  $^{4}$Robotics Institute Germany
}

\usepackage{graphicx}
\usepackage{booktabs}
\usepackage[accsupp]{axessibility}
\usepackage{amsmath}
\usepackage{amssymb}
\usepackage{bm}
\usepackage{multirow}
\usepackage{pifont}
\usepackage{subcaption}
\usepackage{array}
\usepackage{colortbl}
\usepackage{algorithm}
\usepackage{algorithmic}
\usepackage{float}
\usepackage{enumitem}
\usepackage{tcolorbox}
\usepackage{cleveref}
\crefname{figure}{Fig.}{Figs.}
\Crefname{figure}{Fig.}{Figs.}
\crefname{table}{Table}{Tables}
\Crefname{table}{Table}{Tables}
\crefname{section}{Sec.}{Secs.}
\Crefname{section}{Sec.}{Secs.}
\crefname{subsection}{Sec.}{Secs.}
\Crefname{subsection}{Sec.}{Secs.}
\crefname{equation}{Eq.}{Eqs.}
\Crefname{equation}{Eq.}{Eqs.}
\tcbuselibrary{breakable}
\newcolumntype{C}[1]{>{\centering\arraybackslash}p{#1}}

\newcommand{\ours}{BiCICLe}

\DeclareRobustCommand{\myparagraph}[1]{\noindent\textbf{#1}}

\newcommand{\cmark}{\ding{51}}
\newcommand{\xmark}{\ding{55}}

\begin{document}
\maketitle

\begin{abstract}
Large Language Models (LLMs) have emerged as powerful reasoning engines for embodied control. In particular, In-Context Learning (ICL) enables off-the-shelf, text-only LLMs to predict robot actions without any task-specific training while preserving their generalization capabilities. Applying ICL to bimanual manipulation remains challenging as the high-dimensional joint action space and tight inter-arm coordination constraints rapidly overwhelm standard context windows.
To address this, we introduce \textbf{\ours{}} (\textbf{Bi}manual \textbf{C}oordinated \textbf{I}n-\textbf{C}ontext \textbf{Le}arning), the first framework that enables standard LLMs to perform few-shot bimanual manipulation without fine-tuning. \ours{} frames bimanual control as a multi-agent leader-follower problem, decoupling the action space into sequential, conditioned single-arm predictions. Evaluated on 13 tasks from the TWIN benchmark, \ours{} achieves 70.5\% average success rate, outperforming the best training-free baseline by 6.1 percentage points and surpassing most supervised methods. We also demonstrate superior real-world performance on 3 tasks without hardware-specific retraining. The project page is available at \url{https://alesspalma.github.io/bicicle}.

\end{abstract}

\section{Introduction}
\label{sec:intro}

Bimanual manipulation is a cornerstone capability for general-purpose robotic systems. Tasks such as lifting a tray or unscrewing a bottle cap require two arms to synchronize positions, orientations, and forces. This coordination is harder than single-arm control: errors in one arm force the other to compensate, the joint action space grows rapidly, and strict temporal synchronization is required~\cite{lee2015learning,chitnis2020efficient,xie2020deep,grannen2023stabilize}. As a result, Imitation Learning (IL) and offline Reinforcement Learning (RL) typically demand large, task-specific datasets to capture these dependencies.

\begin{figure}[h]
  \centering
  \includegraphics[width=.9\linewidth]{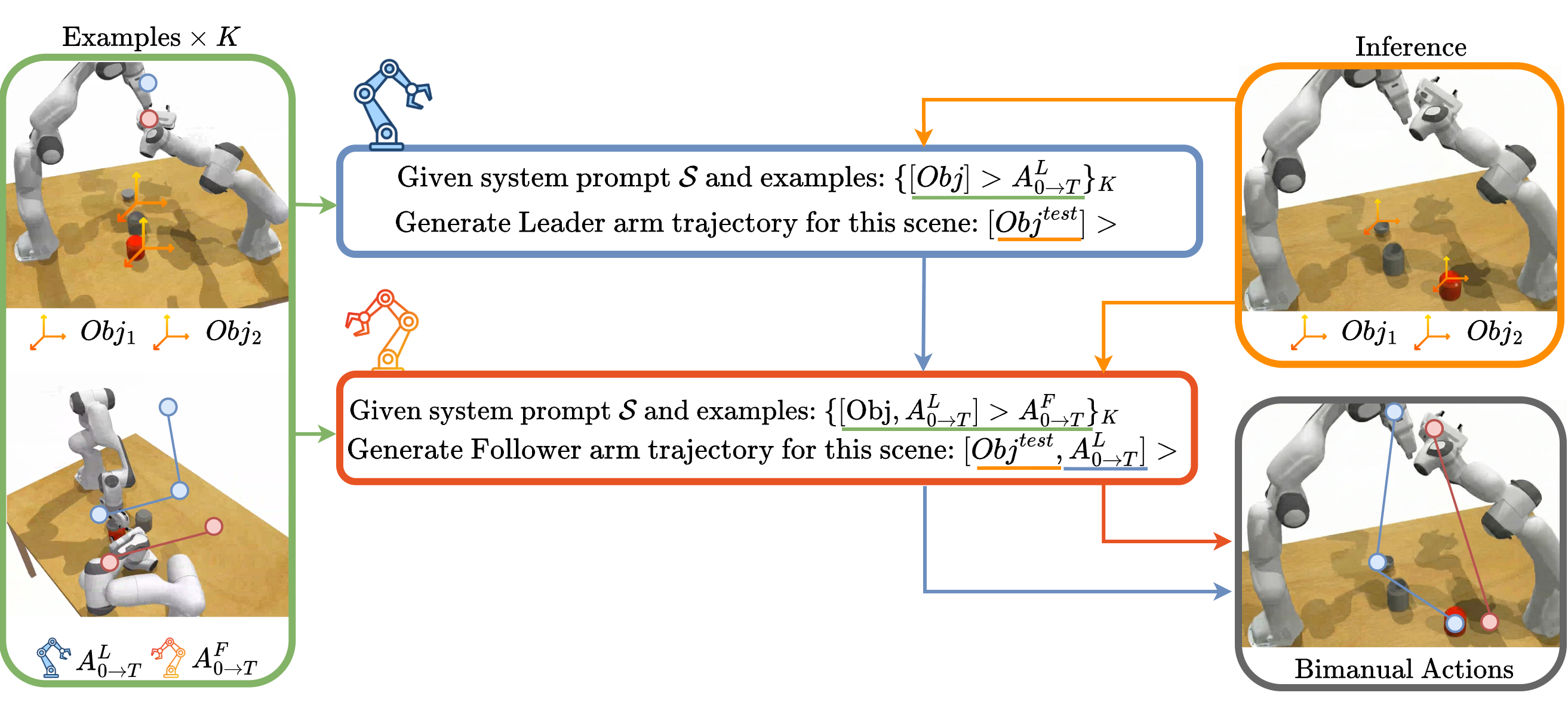}
\caption{\textbf{Overview of the \ours{} Framework.} \emph{(Left)} Bimanual demonstrations are serialized into textual sequences of state observations and actions to construct the in-context prompt. \emph{(Right)} During inference, a leader-follower decomposition enforces inter-arm coordination: the Leader agent predicts its full trajectory first; the Follower agent then predicts its actions conditioned on the Leader's plan. No task-specific training is required.}
  \label{fig:teaser}
\end{figure}

In-Context Learning (ICL) with foundation models offers a compelling alternative: adapting to new robot tasks without any gradient updates. By serializing the scene as text, Large Language Models (LLMs) act as generalist planners, bypassing the need for massive paired image-action datasets. This abstraction trades raw perceptual fidelity for few-shot generalizability, a worthwhile exchange that methods such as RoboPrompt~\cite{roboprompt} and KAT~\cite{kat} have validated for single-arm manipulation. Extending this paradigm to bimanual manipulation, however, remains an open problem. A naive strategy of concatenating both arms' commands into a monolithic joint action sequence inflates token length and increases per-step pattern complexity, causing the LLM to produce incoherent plans. Treating the arms as independent agents is equally untenable: the left arm may attempt a handover before the right arm is positioned to receive. In such contexts, no successful bimanual ICL framework currently exists.

To bridge this gap, we introduce \textbf{Bi}manual \textbf{C}oordinated \textbf{I}n-\textbf{C}ontext \textbf{Le}arning, or \textbf{\ours{}} (\Cref{fig:teaser}), the first multi-agent ICL framework designed for bimanual coordination. Rather than treating bimanual control as a monolithic prediction task, \ours{} adopts a leader-follower architecture 
instantiated as two distinct LLM agents. The Leader predicts its full trajectory from the scene observation. The Follower then predicts its actions conditioned on both the observation and the Leader's complete plan. This factorization enforces inter-arm consistency while reducing the reasoning burden per agent. Our contributions are as follows:

\begin{itemize}
    \item We present \ours{}, the first multi-agent ICL framework that enables 
    LLMs to solve bimanual manipulation tasks through structured leader-follower 
    decomposition, without any task-specific training;
    \item We validate \ours{} on 13 tasks from the TWIN benchmark, where it achieves 70.5\% success rate, outperforming the 
    best training-free baseline by 6.1 percentage points and surpassing most supervised methods. We further demonstrate out-of-distribution generalization on two new tasks;
    \item We deploy \ours{} on a real-world bimanual platform, where it achieves a success rate of 53.3\% across 3 tasks, surpassing the other baselines and showcasing its practical applicability and robustness to real-world complexities.
\end{itemize}


\section{Related Work}
\label{sec:related}

\myparagraph{Robot Manipulation with Language Models.}
LLMs serve robot control in two broad paradigms. High-level planners decompose instructions into skill sequences~\cite{saycan2022arxiv,huang2022language,liang2023code,singh2023progprompt,vemprala2024chatgpt}: SayCan~\cite{saycan2022arxiv} grounds outputs in robot affordances, Code as Policies~\cite{liang2023code} generates executable code, and CaP-X~\cite{fu2026capx} benchmarks and improves code-as-policy agents through execution feedback and skill synthesis. These rely on skill libraries and program execution loops rather than directly producing keyposes from demonstrations. Vision-Language-Action (VLA) models instead fine-tune on robot datasets for end-to-end policies~\cite{brohan2023rt2,kim2024openvla,BlackK-RSS-25}: RT-2~\cite{brohan2023rt2} outputs discretized actions, while $\pi_0$~\cite{BlackK-RSS-25} uses flow matching on a pre-trained backbone. ICL-based approaches instead leverage frozen LLMs: RoboPrompt~\cite{roboprompt} serializes demonstrations as text for pattern completion, and KAT~\cite{kat} uses DINO-ViT keypoint correspondences~\cite{amir2021deep} as observations. Both are limited to single-arm manipulation; to our knowledge ICL has not been successfully adapted to bimanual manipulation, a gap we address.

\myparagraph{Bimanual Robot Manipulation.}
Classical approaches rely on hand-crafted coordination such as master-slave control or relative motion primitives~\cite{smith2012dual,koga1994multi}. Learning-based methods train policies from demonstrations via imitation learning~\cite{xie2020deep,chitnis2020efficient,grannen2023stabilize,lee2024bimact} or reinforcement learning~\cite{chen2022towards}; ACT~\cite{zhao2023act} trains a transformer action-chunking policy from teleoperated demonstrations. More recently, $\pi_0$~\cite{BlackK-RSS-25} achieves bimanual dexterous manipulation by fine-tuning a VLM with flow matching on cross-embodiment data, RoboVLMs~\cite{li2024robovlms} unifies VLM-to-policy fine-tuning, and TwinVLA~\cite{im2026twinvla} composes twin single-arm VLAs for data-efficient bimanual control. All require extensive training on large datasets; in contrast, our approach needs no task-specific training or gradient updates, operating purely through ICL from a few demonstrations.

\myparagraph{In-Context Learning.}
ICL~\cite{brown2020language} lets LLMs learn from prompt examples without parameter updates, with chain-of-thought~\cite{wei2022chain} and self-consistency~\cite{wang2023selfconsistency} adding structured reasoning. In robotics, ICL has aided LLM planning~\cite{huang2022language,liang2023code}, and recent work applies it to continuous control~\cite{sridhar2025ricl,shah2025mimicdroid}; visual ICL, however, remains challenging, as the vision backbone bottlenecks VLA-style approaches~\cite{zhang2026vlm4vla}. Predicting intermediate keyposes from textual observations has instead proven effective for single-arm manipulation~\cite{roboprompt,kat}. We build on this, extending ICL to bimanual manipulation through a leader-follower decomposition as a structured prompting strategy.

\myparagraph{Multi-Agent Coordination with LLMs.}
Multi-agent LLM coordination has been applied to multi-robot planning~\cite{zhang2023building,mandi2024roco}, where agents exchange task-level information before execution. RoCo~\cite{mandi2024roco} assigns an LLM agent per robot, lets them discuss strategy, and validates the resulting plans and waypoints (parsing, task-constraint, IK, and collision checks) before a centralized planner executes them. \ours{} targets a different interface: rather than natural-language deliberation, explicit sub-task allocation, or feedback-driven task-and-motion planning, it uses in-context demonstrations to directly predict discretized bimanual keypose sequences, with a leader-follower conditioning channel that preserves inter-arm coordination while keeping each prediction single-arm.
\section{Method}
\label{sec:method}


We formulate bimanual manipulation as the coordinated plan of two LLM-powered agents. Each arm operates in a 7-dimensional end-effector action space: an SE(3) pose plus a binary gripper command, discretized into $\mathbb{Z}^{7}$. This yields a joint bimanual space $\mathbb{Z}^{14}$ at each keyframe. We first describe the action and observation representations (\cref{sec:formulation}), then present \ours{}, a leader-follower agentic architecture (\cref{sec:leader_follower}).

\subsection{Problem Formulation and Representations}
\label{sec:formulation}

\myparagraph{Task setup.}
We consider a bimanual Franka Panda robot operating in the CoppeliaSim~\cite{coppeliasim} simulation environment with the TWIN benchmark~\cite{twin}. In each episode, the robot observes the scene through six RGB-D cameras (left/right over-shoulder, overhead, left/right wrist, and front). The episode terminates after executing a successful sequence of $K$ keyframe actions for both arms.

\myparagraph{Action discretization.}
Following previous works~\cite{roboprompt, twin, shridhar2022peract}, actions are discretized into integer tokens. The workspace is bounded by a 3D box $\mathcal{B} = [-0.3, -0.5, 0.6] \times [0.7, 0.5, 1.6]$ (meters) and discretized into a $100^3$ voxel grid. Each end-effector position $\mathbf{p} \in \mathbb{R}^3$ is mapped to a voxel index $(v_x, v_y, v_z) \in \{0, \ldots, 99\}^3$ via:
\begin{equation}
v_i = \left\lfloor \frac{p_i - \mathcal{B}^{\min}_i}{\mathcal{B}^{\max}_i - \mathcal{B}^{\min}_i} \cdot 99 \right\rfloor, \quad i \in \{x, y, z\},
\label{eq:voxelization}
\end{equation}
where $\mathcal{B}^{\min}_i$ and $\mathcal{B}^{\max}_i$ are the bounds along axis $i$. End-effector orientations, represented as quaternions, are converted to Euler angles and discretized into bins of $5^{\circ}$, yielding rotation indices $(r_x, r_y, r_z) \in \{0, \ldots, 71\}^3$. The gripper state is binarized as $g \in \{0, 1\}$ (closed/open). A single-arm action is thus a 7-tuple $\mathbf{a} = (v_x, v_y, v_z, r_x, r_y, r_z, g)$, and a bimanual action is the concatenation $\mathbf{a}^{\text{bi}} = [\mathbf{a}^R, \mathbf{a}^L] \in \mathbb{Z}^{14}$.

\myparagraph{Observation representation.}
At each timestep, the observation consists of the 3D positions of task-relevant objects. Object positions are computed by fusing segmentation masks and point clouds from all six cameras: for each object, masked point clouds are merged across views, downsampled using voxel grid filtering~\cite{open3d}, and the centroid is discretized to voxel coordinates using \cref{eq:voxelization}. We ablate this point-cloud extraction choice in \Cref{sec:ablation_pointcloud}. The observation is represented as a dictionary mapping object names to discretized positions as 
    $\mathbf{o} = \{\texttt{obj}_1: [x_1, y_1, z_1], \ldots, \texttt{obj}_M: [x_M, y_M, z_M]\}.$
This text-based representation is compact yet informative, encoding the spatial relationships between objects and the robot's end-effectors that are critical for bimanual coordination. In contrast to~\cite{roboprompt}, we found that including object orientations in the observation degrades performance on the simulation benchmark (see \Cref{sec:ablation_rotations}).

\myparagraph{In-context demonstrations.}
A demonstration $\mathcal{D}_i$ consists of an initial scene observation $\mathbf{o}_i$ paired with $K_i$ bimanual keyframe actions $\mathbf{A}_i = [\mathbf{a}_{i,1}^{\text{bi}}, \ldots, \mathbf{a}_{i,K_i}^{\text{bi}}]$. Following previous work~\cite{twin}, keyframes are identified via a heuristic that detects gripper state changes in either arm, zero joint velocities, and episode termination. $N = 10$ demonstrations are serialized as text and prepended to the test observation $\mathbf{o}_{\text{test}}$ to form the prompt as
    $``\mathbf{o}_1 \texttt{\char`\>} \mathbf{A}_1\texttt{, } \ldots\texttt{, } \mathbf{o}_N \texttt{\char`\>} \mathbf{A}_N\texttt{, } \mathbf{o}_{\text{test}} \texttt{\char`\>}"$
The exact prompt templates used by \ours{} are reported in \Cref{sec:supp_prompts}.

\subsection{\ours{}: A Leader-Follower LLM Architecture}
\label{sec:leader_follower}

\ours{} provides inter-arm coordination without increasing the per-agent action space or the context length of each LLM call. Rather than predicting the full sequence of bimanual actions $\mathbf{a}^{\text{bi}} \in \mathbb{Z}^{14}$ in a single agent call, we factor the problem into two sequential $\mathbb{Z}^{7}$ predictions linked by explicit conditioning.

\myparagraph{Phase 1: Leader prediction.}
Once an arm is designated as the leader, the bimanual demonstrations are stripped to single-arm format: only the leader arm's actions are retained. The leader agent receives a system prompt, followed by the single-arm ICL demonstrations and the test observation:
\begin{equation}
    \texttt{prompt}^L = \mathbf{o}_1 \texttt{\char`\>} \mathbf{A}_1^L\texttt{, } \ldots\texttt{, } \mathbf{o}_N \texttt{\char`\>} \mathbf{A}_N^L\texttt{, } \mathbf{o}_{\text{test}} \texttt{\char`\>}
\label{eq:leader_prompt}
\end{equation}
where $\mathbf{A}_i^L = [\mathbf{a}_{i,1}^L, \ldots, \mathbf{a}_{i,K_i}^L]$ denotes the leader arm's actions from demonstration $i$. Note that we omit any textual description of the task; the agent automatically infers the objective by recognizing patterns within the demonstrations. The leader agent generates the predicted leader trajectory $\hat{\mathbf{A}}^L = [\hat{\mathbf{a}}_1^L, \ldots, \hat{\mathbf{a}}_{\hat{K}_L}^L]$.

\myparagraph{Phase 2: Follower prediction.}
The follower agent predicts its actions \emph{conditioned on the leader's predicted trajectory}. To achieve this, the in-context demonstrations are restructured: the leader arm's ground-truth actions are embedded directly into the observation dictionary as an additional entry, creating an augmented observation:
\begin{equation}
    \tilde{\mathbf{o}}_i = \{\texttt{obj}_1: [\cdot], \ldots, \texttt{obj}_M: [\cdot], \texttt{leader\_arm}: [\mathbf{a}_{i,1}^L, \ldots, \mathbf{a}_{i,K_i}^L]\},
\label{eq:follower_obs}
\end{equation}
and the demonstration actions are replaced with the follower's single-arm actions $\mathbf{A}_i^F$. Then, the leader's \emph{predicted} actions $\hat{\mathbf{A}}^L$ are inserted into the test observation to form the follower's prompt:
\begin{equation}
    \texttt{prompt}^F = \tilde{\mathbf{o}}_1 \texttt{\char`\>} \mathbf{A}_1^F\texttt{, } \ldots\texttt{, } \tilde{\mathbf{o}}_N \texttt{\char`\>} \mathbf{A}_N^F\texttt{, } \tilde{\mathbf{o}}_{\text{test}} \texttt{\char`\>}
\label{eq:follower_prompt}
\end{equation}
where $\tilde{\mathbf{o}}_{\text{test}} = \{\texttt{objects}, \texttt{leader\_arm}: \hat{\mathbf{A}}^L\}$. The follower agent generates the predicted follower trajectory $\hat{\mathbf{A}}^F = [\hat{\mathbf{a}}_1^F, \ldots, \hat{\mathbf{a}}_{\hat{K}_F}^F]$.

\myparagraph{Action composition.}
The leader and follower trajectories are combined into the bimanual action sequence, mapping back to the right and left arms according to the leader assignment. If the two sequences differ in length, the shorter is extended by repeating its last action:
\begin{equation}
    \hat{\mathbf{a}}_k^{\text{bi}} = [\hat{\mathbf{a}}_k^R, \hat{\mathbf{a}}_k^L], \quad k = 1, \ldots, \max(\hat{K}_L, \hat{K}_F).
\label{eq:composition}
\end{equation}

Following \cite{roboprompt}, \ours{} executes predicted trajectories one keyframe at a time. This chunked closed-loop policy reobserves and replans when the queue empties before success, within a total budget of 25 environment steps per episode.

\section{Experiments}
\label{sec:experiments}

We first evaluate \ours{} in simulation, covering the benchmark setup, quantitative results, and qualitative failure modes. We then transfer the same pose-based ICL interface to a physical bimanual Franka Panda system.

\subsection{Simulation Experimental Setup}
\label{sec:setup}

We evaluate our approach on the TWIN benchmark~\cite{twin}, a bimanual extension of RLBench~\cite{james2020rlbench} that provides 13 bimanual manipulation tasks with varying sequence lengths and coordination requirements. All experiments utilize CoppeliaSim~\cite{coppeliasim} with a simulated bimanual Franka Panda robot and six RGB-D cameras at a $128 \times 128$ resolution. For each task, 100 training demonstrations and 100 test demonstrations are generated using the oracular CoppeliaSim motion planner. Keyframes are extracted using the bimanual heuristic (\cref{sec:formulation}) and grouped into batches of $N=10$ ICL demonstrations. Unless otherwise specified, all training-free methods in \Cref{tab:main_results} utilize GPT-5-mini\footnote{We use the \texttt{gpt-5-mini-2025-08-07} snapshot via the OpenAI Chat Completions endpoint.}~\cite{singh2025openaigpt5card}, and each task is evaluated over 3 seeds $\times$ 100 episodes. With Qwen~2.5 7B, \ours{} retains its advantage ($55.6\%$ vs.\ $51.9\%$ for RoboPrompt-SA; \Cref{sec:ablation_backbone}).

\myparagraph{ICL and numerical baselines.}
We denote monolithic approaches that predict joint bimanual actions in a single agent call as SA (Single Agent), and approaches using per-arm independent calls with no information sharing as DA (Dual Agent).
RoboPrompt-SA / RoboPrompt-DA: adapted from RoboPrompt~\cite{roboprompt}, using text-based object positions.
KAT-SA / KAT-DA: adapted from KAT~\cite{kat}, these methods use DINO-ViT keypoint correspondences~\cite{amir2021deep,caron2021emerging}.
VLM-LF: A leader-follower variant inspired by few-shot VLM prompting~\cite{alayrac2022flamingo}, using interleaved RGB+depth front-camera images and action-token demonstrations.
XGBoost~\cite{chen2016xgboost}: a numerical baseline trained only on each sampled ten-demo context, predicting bimanual keyframes from object centroids and normalized trajectory phase.

\myparagraph{Supervised methods.}
All supervised results in \Cref{tab:main_results} are reported from 3DFA~\cite{ze20253dfa}. The reported settings use 100 training demonstrations and 100 evaluation episodes per task, but differ in observations, action interfaces, training regimes, checkpoint selection, and task coverage.

ACT~\cite{zhao2023act}, RVT-LF, PerAct-LF, and PerAct$^2$~\cite{twin} are trained separately per task, with the best validation checkpoint evaluated on a separate test set. Images have a $256\times256$ resolution, compared with our $128\times128$. ACT uses front and two wrist views to predict joint-position chunks. The other three use five RGB-D views (front, two shoulders, and two wrists) to predict end-effector keyposes executed by a motion planner. RVT-LF and PerAct-LF predict the arms sequentially, whereas PerAct$^2$ predicts them jointly. PerAct-LF and PerAct$^2$ use the same $100^3$ grid and $5^\circ$ rotation bins as \ours{}; RVT-LF predicts positions from rendered-view heatmaps rather than voxels, retaining $5^\circ$ rotation bins.

DP3~\cite{ze2024dp3} and KStarDiffuser~\cite{kstardiffuser2024} are evaluated on five tasks, with results averaged over three seeds. DP3 uses point clouds to predict joint angles through diffusion; KStarDiffuser predicts continuous end-effector keyposes regularized by robot kinematics. PPI~\cite{ppi2024} covers seven tasks using regenerated training data and reports averages over three checkpoints. It combines multi-view 3D features, gripper-keypose prediction, and object-pointflow supervision to generate continuous actions. These continuous outputs do not use our voxel and rotation bins.

AnyBimanual~\cite{anybimanual2024}, $\pi_0$-keypose~\cite{ze20253dfa}, and 3DFA use multi-task training. AnyBimanual adapts pretrained unimanual PerAct~\cite{shridhar2022peract} policies on 12 tasks, using six $256\times256$ camera views and a $100^3$ voxel grid. For $\pi_0$-keypose, pretrained $\pi_0$ is fine-tuned to predict end-effector keyposes rather than its native joint-action chunks. Both $\pi_0$-keypose and 3DFA use front and two wrist cameras at $256\times256$, continuous keypose outputs, and final checkpoints shared across all 13 tasks.

These supervised policies learn from visual observations, while \ours{} receives object centroids estimated from RGB-D point clouds with ground-truth segmentation masks in simulation and performs no gradient updates, using $N{=}10$ in-context examples. The supervised rows therefore provide literature reference points rather than matched comparisons of perception or sample efficiency.

\subsection{Simulation Results} 
\label{sec:main_results}

\definecolor{supgray}{gray}{0.88}
\providecommand{\twintabmeanpad}[1]{\ifdim #1pt<10pt \phantom{00}\else\ifdim #1pt<100pt \phantom{0}\fi\fi}
\providecommand{\twintabstdpad}[1]{\ifdim #1pt<10pt \phantom{0}\fi}
\providecommand{\twintabmean}[1]{\twintabmeanpad{#1}#1}
\providecommand{\twintabstd}[1]{\twintabstdpad{#1}#1}
\providecommand{\twintabnum}[1]{\ensuremath{\twintabmean{#1}}}
\providecommand{\twintabbestnum}[1]{\ensuremath{\twintabmeanpad{#1}\bm{#1}}}
\providecommand{\twintabsecondnum}[1]{\ensuremath{\twintabmeanpad{#1}\underline{#1}}}
\providecommand{\twintabres}[2]{\ensuremath{\twintabmean{#1}_{\pm\twintabstd{#2}}}}
\providecommand{\twintabbestres}[2]{\ensuremath{\twintabmeanpad{#1}\bm{#1}_{\bm{\pm}\twintabstdpad{#2}\bm{#2}}}}
\providecommand{\twintabsecondres}[2]{\ensuremath{\twintabmeanpad{#1}\underline{#1_{\pm\twintabstd{#2}}}}}
\begin{table*}[h]
\caption{\textbf{Success rates (\%) on the TWIN benchmark.} Few-shot methods: mean $\pm$ std over 3 runs $\times$ 100 episodes. \textbf{Bold}: best among few-shot methods per task. \underline{Underline}: second best. \colorbox{supgray}{Gray rows:} supervised methods; results reported from~\cite{ze20253dfa}, {--} denotes unavailable results.}
\label{tab:main_results}
\centering
\resizebox{\textwidth}{!}{%
\begin{tabular}{l|rrrrrrrrrrrrr|r}
\toprule
\textbf{Method} & \rotatebox{70}{Push Box} & \rotatebox{70}{Dual Buttons} & \rotatebox{70}{Bottle Fridge} & \rotatebox{70}{Handover} & \rotatebox{70}{Handover Easy} & \rotatebox{70}{Lift Ball} & \rotatebox{70}{Lift Tray} & \rotatebox{70}{Pick Laptop} & \rotatebox{70}{Pick Plate} & \rotatebox{70}{Straighten Rope} & \rotatebox{70}{Sweep Dustpan} & \rotatebox{70}{Tray Oven} & \rotatebox{70}{Item Drawer} & \textbf{Avg.} \\
\midrule
\rowcolor{supgray}\multicolumn{15}{l}{\textit{Supervised baselines (results from~\cite{ze20253dfa})}} \\
\rowcolor{supgray} ACT~\cite{zhao2023act} & \twintabnum{0.0} & \twintabnum{4.0} & \twintabnum{0.0} & \twintabnum{0.0} & \twintabnum{0.0} & \twintabnum{36.0} & \twintabnum{6.0} & \twintabnum{0.0} & \twintabnum{0.0} & \twintabnum{16.0} & \twintabnum{0.0} & \twintabnum{2.0} & \twintabnum{13.0} & \twintabnum{5.9} \\
\rowcolor{supgray} RVT-LF~\cite{twin} & \twintabnum{52.0} & \twintabnum{39.0} & \twintabnum{0.0} & \twintabnum{0.0} & \twintabnum{0.0} & \twintabnum{17.0} & \twintabnum{6.0} & \twintabnum{3.0} & \twintabnum{3.0} & \twintabnum{3.0} & \twintabnum{0.0} & \twintabnum{3.0} & \twintabnum{10.0} & \twintabnum{10.5} \\
\rowcolor{supgray} PerAct-LF~\cite{twin} & \twintabnum{57.0} & \twintabnum{10.0} & \twintabnum{0.0} & \twintabnum{0.0} & \twintabnum{9.0} & \twintabnum{40.0} & \twintabnum{14.0} & \twintabnum{11.0} & \twintabnum{2.0} & \twintabnum{21.0} & \twintabnum{28.0} & \twintabnum{8.0} & \twintabnum{27.0} & \twintabnum{17.5} \\
\rowcolor{supgray} PerAct$^2$~\cite{twin} & \twintabnum{6.0} & \twintabnum{47.0} & \twintabnum{3.0} & \twintabnum{11.0} & \twintabnum{41.0} & \twintabnum{50.0} & \twintabnum{1.0} & \twintabnum{12.0} & \twintabnum{4.0} & \twintabnum{24.0} & \twintabnum{0.0} & \twintabnum{9.0} & \twintabnum{10.0} & \twintabnum{16.8} \\
\rowcolor{supgray} DP3~\cite{ze2024dp3} & \twintabnum{56.0} & {--} & {--} & {--} & \twintabnum{0.0} & \twintabnum{64.0} & {--} & \twintabnum{6.3} & {--} & {--} & \twintabnum{1.7} & {--} & {--} & {--} \\
\rowcolor{supgray} KStarDiffuser~\cite{kstardiffuser2024} & \twintabnum{83.0} & {--} & {--} & {--} & \twintabnum{27.0} & \twintabnum{98.7} & {--} & \twintabnum{43.7} & {--} & {--} & \twintabnum{89.0} & {--} & {--} & {--} \\
\rowcolor{supgray} PPI~\cite{ppi2024} & \twintabnum{96.7} & {--} & {--} & {--} & \twintabnum{62.7} & \twintabnum{89.3} & \twintabnum{92.0} & \twintabnum{46.3} & {--} & {--} & \twintabnum{98.7} & {--} & \twintabnum{79.7} & {--} \\
\rowcolor{supgray} AnyBimanual~\cite{anybimanual2024} & \twintabnum{46.0} & \twintabnum{73.0} & \twintabnum{26.0} & \twintabnum{15.0} & \twintabnum{44.0} & \twintabnum{36.0} & \twintabnum{14.0} & \twintabnum{7.0} & \twintabnum{8.0} & \twintabnum{24.0} & \twintabnum{67.0} & \twintabnum{24.0} & {--} & \twintabnum{32.0} \\
\rowcolor{supgray} $\pi_0$-keypose~\cite{ze20253dfa} & \twintabnum{93.0} & \twintabnum{38.0} & \twintabnum{22.0} & \twintabnum{2.0} & \twintabnum{59.0} & \twintabnum{97.0} & \twintabnum{72.0} & \twintabnum{27.0} & \twintabnum{41.0} & \twintabnum{7.0} & \twintabnum{2.0} & \twintabnum{68.0} & \twintabnum{40.0} & \twintabnum{43.7} \\
\rowcolor{supgray} 3DFA~\cite{ze20253dfa} & \twintabres{92.7}{0.5} & \twintabres{92.7}{1.9} & \twintabres{89.3}{1.9} & \twintabres{89.0}{7.1} & \twintabres{96.0}{5.7} & \twintabres{99.7}{0.5} & \twintabres{94.7}{0.5} & \twintabres{74.0}{9.0} & \twintabres{69.7}{12.6} & \twintabres{40.7}{1.9} & \twintabres{99.3}{0.5} & \twintabres{94.7}{1.9} & \twintabres{93.0}{2.8} & \twintabnum{85.1} \\
\midrule
\multicolumn{15}{l}{\textit{Few-shot baselines (ten demonstrations)}} \\
XGBoost~\cite{chen2016xgboost} & \twintabsecondres{97.3}{1.7} & \twintabsecondres{86.7}{2.5} & \twintabres{42.7}{7.1} & \twintabres{37.0}{3.3} & \twintabres{30.7}{4.0} & \twintabsecondres{80.7}{3.9} & \twintabres{48.7}{2.9} & \twintabres{11.0}{2.2} & \twintabres{33.0}{2.9} & \twintabres{12.0}{0.8} & \twintabsecondres{95.0}{1.4} & \twintabres{18.7}{2.5} & \twintabres{45.0}{0.8} & \twintabnum{49.1} \\
VLM-LF & \twintabres{59.0}{7.0} & \twintabres{6.3}{1.2} & \twintabres{7.0}{1.6} & \twintabres{6.7}{0.5} & \twintabres{8.3}{1.2} & \twintabres{11.3}{1.2} & \twintabres{8.0}{0.8} & \twintabres{5.7}{0.9} & \twintabres{7.7}{1.2} & \twintabres{3.0}{0.0} & \twintabres{31.0}{4.2} & \twintabres{10.0}{2.4} & \twintabres{10.0}{0.8} & \twintabnum{13.4} \\
KAT-SA~\cite{kat} & \twintabres{60.0}{2.0} & \twintabres{4.3}{2.5} & \twintabres{11.7}{1.5} & \twintabres{4.0}{1.0} & \twintabres{33.7}{2.5} & \twintabres{42.3}{4.5} & \twintabres{17.0}{1.0} & \twintabres{2.3}{0.5} & \twintabres{3.7}{0.5} & \twintabres{2.7}{0.5} & \twintabres{55.3}{0.5} & \twintabres{10.0}{1.0} & \twintabres{16.3}{3.5} & \twintabnum{20.3} \\
KAT-DA~\cite{kat} & \twintabres{50.7}{5.0} & \twintabres{3.0}{2.0} & \twintabres{17.3}{2.0} & \twintabres{4.0}{1.0} & \twintabres{30.7}{9.5} & \twintabres{19.0}{2.0} & \twintabres{11.3}{0.5} & \twintabres{0.3}{0.5} & \twintabres{2.3}{0.5} & \twintabres{1.7}{0.5} & \twintabres{74.3}{1.5} & \twintabres{6.0}{0.0} & \twintabres{6.0}{3.0} & \twintabnum{17.4} \\
RoboPrompt-SA~\cite{roboprompt} & \twintabres{94.0}{0.8} & \twintabbestres{100.0}{0.0} & \twintabbestres{82.0}{3.6} & \twintabsecondres{83.7}{2.6} & \twintabsecondres{64.7}{2.1} & \twintabres{78.7}{5.2} & \twintabres{58.7}{4.0} & \twintabres{18.0}{3.0} & \twintabres{44.0}{1.6} & \twintabres{11.7}{1.7} & \twintabres{89.3}{1.9} & \twintabres{26.3}{2.3} & \twintabres{37.3}{3.1} & \twintabnum{60.6} \\
RoboPrompt-DA~\cite{roboprompt} & \twintabres{83.0}{2.8} & \twintabbestres{100.0}{0.0} & \twintabres{80.0}{2.9} & \twintabres{82.7}{3.7} & \twintabres{63.3}{3.4} & \twintabres{69.3}{0.5} & \twintabsecondres{79.3}{2.6} & \twintabsecondres{24.3}{2.1} & \twintabsecondres{61.0}{2.0} & \twintabsecondres{23.3}{2.3} & \twintabres{93.7}{0.9} & \twintabsecondres{30.0}{1.7} & \twintabbestres{47.0}{2.9} & \twintabsecondnum{64.4} \\
\midrule

\textbf{\ours{}} (ours) & \twintabbestres{99.0}{1.0} & \twintabbestres{100.0}{0.0} & \twintabsecondres{80.3}{2.9} & \twintabbestres{94.3}{1.5} & \twintabbestres{68.0}{2.6} & \twintabbestres{83.7}{2.3} & \twintabbestres{83.0}{2.9} & \twintabbestres{29.0}{2.0} & \twintabbestres{65.3}{2.4} & \twintabbestres{34.3}{2.1} & \twintabbestres{97.3}{1.2} & \twintabbestres{36.0}{1.4} & \twintabsecondres{46.7}{2.4} & \twintabbestnum{70.5} \\
\bottomrule
\end{tabular}%
}
\end{table*}

\Cref{tab:main_results} reports success rates across all 13 tasks. \ours{} achieves the best average performance among training-free ICL methods (70.5\%), outperforming the strongest baseline by 6.1 percentage points and improving on most tasks. Context-fitted XGBoost reaches $49.1\%$, $21.4$ points below \ours{}. It is competitive on Push Box, Lift Ball, and Sweep Dustpan, but trails \ours{} markedly on Handover ($37.0\%$ vs.\ $94.3\%$) and Pick Plate ($33.0\%$ vs.\ $65.3\%$). The supervised rows serve as literature reference points; as discussed above, their setups can differ in camera count, observation representation, action interface, and amount of supervision. Additional qualitative examples are provided in \Cref{sec:supp_qualitatives}.

\myparagraph{Supervised vs.\ ICL.}
Supervised methods (3DFA~\cite{ze20253dfa}: 85.1\%) can achieve higher overall success after task-specific training. Our goal is to show that a training-free policy can be competitive: \ours{} exceeds the reported success rates of ACT~\cite{zhao2023act} ($5.9\%$), PerAct$^2$~\cite{twin} ($16.8\%$), $\pi_0$-keypose~\cite{ze20253dfa} ($43.7\%$), and AnyBimanual~\cite{anybimanual2024} ($32.0\%$), as well as the reported SOTA on Push Box, Dual Buttons, and Handover. These comparisons retain the protocol differences in \Cref{sec:setup}. The LLM predicts sparse keyposes from semantic object coordinates rather than dense motor commands, turning manipulation into sequence completion over spatial tokens~\cite{pmlr-v229-mirchandani23a}.

\myparagraph{SA vs.\ DA: a taxonomy-based analysis.}
We analyze RoboPrompt-SA (joint 14D prediction) and RoboPrompt-DA (two independent 7D predictions) through the bimanual taxonomy of Krebs and Asfour~\cite{krebs2022tax}. Their relative performance follows the expected coupling structure. On simple tightly coupled \emph{symmetric} tasks, SA's joint prediction captures correlated motion, beating DA on Push Box (+11.0) and Lift Ball (+9.4); this advantage reverses when per-arm precision dominates, e.g.\ Lift Tray (DA +20.6). On tightly coupled \emph{asymmetric} tasks, where the arms play distinct roles, DA is usually stronger (Pick Plate +17.0, Straighten Rope +11.6, Pick Laptop +6.3). On \emph{loosely coupled} tasks the gap is small ($\leq$9.7), as the arms act largely independently.

\myparagraph{\ours{}: merging the best of both worlds.}
This exposes the central trade-off: SA models inter-arm correlation but suffers from high-dimensional prediction, while DA reduces dimensionality but discards coordination. \ours{} keeps DA's 7D per-arm format while reintroducing coordination by conditioning the follower on the leader trajectory. DA improves over SA by 3.8 points on average; \ours{} adds 6.1 over DA, showing that explicit conditioning helps beyond factorization alone (\Cref{sec:ablation_sequential}). The largest gains arise when coordination and low dimensionality are both needed, as in Straighten Rope (+22.6 over SA, +11.0 over DA). These trends support the leader-follower asymmetry: many bimanual tasks exhibit primary-secondary roles~\cite{krebs2022tax}, and even symmetric tasks can be successfully coordinated by conditioning one arm on the other.

\myparagraph{Observation representations.} The KAT baselines (DINO-ViT keypoint correspondences) drastically underperform text-based baselines, indicating that semantic object identities are far more informative than appearance-based keypoints for ICL, particularly where precise object-relative positioning is critical. Despite accessing RGB+depth images, VLM-LF ($13.4\%$) ranks last, confirming that discretized text observations already capture the spatial information needed for bimanual coordination.

\myparagraph{Ablations and inference cost.}
\Cref{tab:ablation_cost_summary} summarizes the ablation results and inference costs. Sequential Arms retains sequential 7D predictions in a shared chat but removes the leader-follower observation augmentation, reaching $65.0\%$, $5.5$ points below \ours{}, supporting the benefit of structured conditioning. Conversational cross-arm refinement degrades success through gripper-state inversions and coordinate drift. Best-of-N samples and scores five bimanual plans, gaining only $0.7$ points at $8.4\times$ the tokens on Lift Ball. \ours{} improves success over RoboPrompt-DA but incurs higher latency; parallel sampling limits Best-of-N's wall-time increase despite its token overhead. Detailed ablation results and full cost statistics are in \Cref{sec:supp_ablations,sec:supp_latency}.

\begin{table}[htbp]
\centering
\caption{\textbf{Ablations and inference cost.} Success averaged over 13 tasks; mean total tokens and median LLM-call time per episode measured on Lift Ball, including API latency. Bold: highest success; --: not measured.}
\label{tab:ablation_cost_summary}
\begin{tabular}{l|rrr}
\toprule
\textbf{Variant} & Success ($\%$) & Tokens (k) & Time (s) \\
\midrule
RoboPrompt-DA & \twintabnum{64.4} & \twintabnum{17.5} & \twintabnum{43.3} \\
Sequential Arms & \twintabnum{65.0} & -- & -- \\
\ours{} & \twintabnum{70.5} & \twintabnum{21.5} & \twintabnum{125.3} \\
+ Conversation & \twintabnum{57.5} & \twintabnum{73.8} & \twintabnum{278.2} \\
+ Best-of-N & \twintabbestnum{71.2} & \twintabnum{180.7} & \twintabnum{149.7} \\
\bottomrule
\end{tabular}
\end{table}

\begin{figure}[H]
  \centering
  \includegraphics[width=0.99\linewidth]{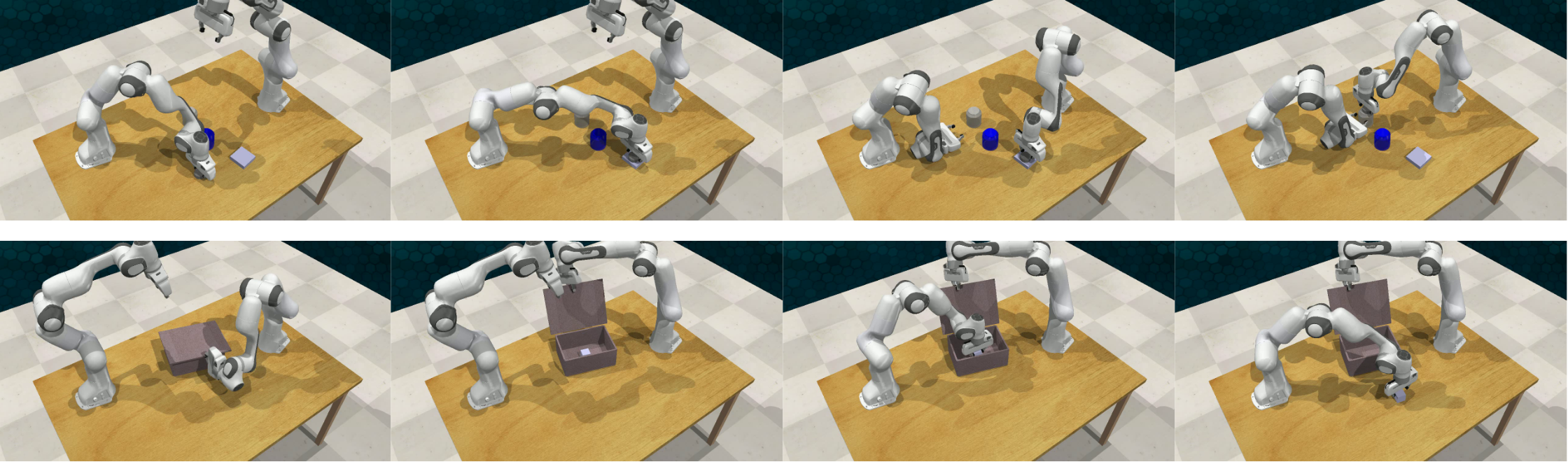}
  \caption{\textbf{New generalization tasks.} Two bimanual tasks designed outside the TWIN benchmark. \emph{(Top)} \textbf{Close Jar}. \emph{(Bottom)} \textbf{Take Item Out of Box}.}
  \label{fig:our_tasks}
\end{figure}

\myparagraph{Generalization to New Tasks.} A key advantage of ICL is generalizing to novel tasks without retraining, since a few test-time demonstrations suffice. We design two bimanual tasks (\cref{fig:our_tasks}) outside the original TWIN benchmark:
\begin{itemize}[nosep,leftmargin=*]
    \item \textbf{Close Jar}: One arm hands over the lid and then holds a jar in place, while the other picks up the lid and places it on top of the target jar.
    \item \textbf{Take Item Out of Box}: One arm lifts the box lid open while the other grasps the item inside and places it on the table.
\end{itemize}

We compare \ours{} against 3DFA fine-tuned on $N=10$ demonstrations per task for 80,000 steps, using full fine-tuning (3DFA-fft) and LoRA~\cite{hu2022lora} (3DFA-LoRA), with all methods evaluated on 100 episodes per task.
\Cref{tab:generalization} reveals a large gap: \ours{} reaches $54.5$\% average success, versus $10.0$\% for 3DFA-fft and $15.0$\% for 3DFA-LoRA. LoRA helps in this low-data setting but stays far below the training-free result. Despite being the TWIN state of the art, 3DFA does not transfer to out-of-distribution tasks without sufficient task-specific data, whereas \ours{} adapts to entirely new tasks through prompt demonstrations alone.

\begingroup
\setlength{\intextsep}{4pt}
\setlength{\abovecaptionskip}{3pt}
\setlength{\belowcaptionskip}{2pt}
\begin{table}[H]
\centering
\caption{\textbf{Generalization.} Success rates ($\%$) on two tasks outside the TWIN benchmark.}
\label{tab:generalization}
\begin{tabular}{l|rr|r}
\toprule
\textbf{Method} & Close Jar & Take Item & \textbf{Avg.} \\
\midrule
3DFA-fft~\cite{ze20253dfa} & \twintabnum{11.0} & \twintabnum{9.0} & \twintabnum{10.0} \\
3DFA-LoRA~\cite{hu2022lora} & \twintabsecondnum{12.0} & \twintabsecondnum{18.0} & \twintabsecondnum{15.0} \\
\ours{} & \twintabbestnum{61.0} & \twintabbestnum{48.0} & \twintabbestnum{54.5} \\
\bottomrule
\end{tabular}
\end{table}
\endgroup

\subsection{Real-World Deployment}
\label{sec:real_world}
We finally validate whether the same pose-based ICL interface transfers beyond simulation by deploying \ours{} on a physical bimanual Franka Panda system. The setup consists of two Franka Panda Research 3 arms and a Stereolabs ZED X camera mounted from an egocentric viewpoint. For each task, we collect 15 kinesthetic demonstrations on the physical robot and sample $N=10$ demonstrations in the prompt at test time, matching the simulation protocol. Object poses are estimated with FoundationPose~\cite{Wen2023FoundationPoseU6}, discretized within workspace bounds adapted to the real robot, and augmented with object yaw, which we found to be reliable in the physical setup. The predicted 6D end-effector poses and gripper states are executed through MoveIt~\cite{coleman2014reducing}.

\begin{figure}[h]
  \centering
  \includegraphics[width=0.99\linewidth]{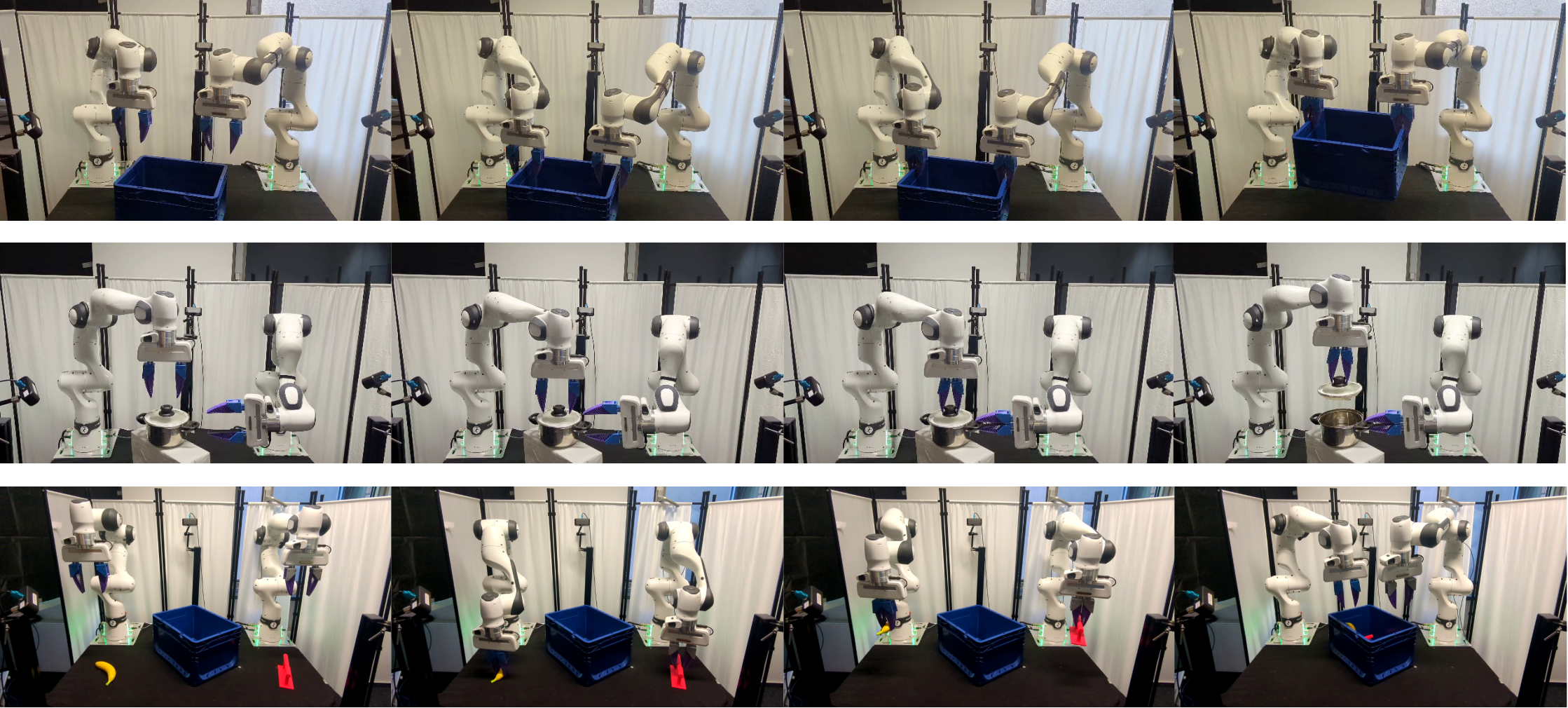}
  \caption{\textbf{Real-world task executions.} \emph{Top row:} \textbf{Lift Box}. \emph{Middle row:} \textbf{Open Pot}. \emph{Bottom row:} \textbf{Cleanup}. Successful episodes completed by \ours{} on a physical bimanual Franka Panda system.}
  \label{fig:real_world}
\end{figure}

We evaluate the three real-world tasks shown in \Cref{fig:real_world} over 10 trials each with varied object locations. \textbf{Lift Box} requires both arms to grasp opposite sides of a box-like container and lift it jointly, testing tight symmetric coordination. \textbf{Open Pot} requires one arm to hold the pot in place while the other grasps and lifts the lid, testing asymmetric role assignment and fine manipulation. \textbf{Cleanup} requires the robot to place both a banana and a handle into a basket, testing independent object manipulation. We compare against two ICL baselines from simulation, RoboPrompt-DA and KAT-DA.

\begingroup
\setlength{\intextsep}{4pt}
\setlength{\abovecaptionskip}{3pt}
\setlength{\belowcaptionskip}{2pt}
\begin{table}[H]
\centering
\caption{\textbf{Real-world deployment.} Success rates ($\%$) over 10 trials per task.}
\label{tab:real_world}
\begin{tabular}{l|rrr|r}
\toprule
\textbf{Method} & Lift Box & Open Pot & Cleanup & \textbf{Avg.} \\
\midrule
KAT-DA~\cite{kat} & \twintabnum{40.0} & \twintabnum{20.0} & \twintabnum{10.0} & \twintabnum{23.3} \\
RoboPrompt-DA~\cite{roboprompt} & \twintabnum{30.0} & \twintabbestnum{40.0} & \twintabnum{30.0} & \twintabnum{33.3} \\
\ours{} & \twintabbestnum{60.0} & \twintabbestnum{40.0} & \twintabbestnum{60.0} & \twintabbestnum{53.3} \\
\bottomrule
\end{tabular}
\end{table}
\endgroup

\Cref{tab:real_world} shows that \ours{} transfers better than the baselines on the completed real-world comparisons, reaching $60.0\%$ on Lift Box, $40.0\%$ on Open Pot, and $60.0\%$ on Cleanup, with the highest gains on the more complex Cleanup task, showing the usefulness of our Leader-Follower approach as compared to a more convoluted and larger context in KAT-DA and RoboPrompt-DA. KAT-DA struggles in the real-world setting due to the inherent limitations of appearance-based keypoint representations: keypoint correspondences are inconsistent across varying object poses, arising from its sensitivity to texture, illumination, reflectance, etc., preventing the model from reliably recovering the spatial relationships critical for bimanual coordination. 
The remaining failures are consistent with the simulation limitations discussed above: grasp predictions can be misaligned under coarse discretization, and small handles or partially occluded objects make the policy sensitive to pose-estimation noise. Overall, the same training-free, demonstration-conditioned policy can execute physically demanding bimanual tasks on real hardware, while perception accuracy and discretization granularity remain the main bottlenecks for further improvement.

\section{Limitations}
Our simulation perception uses ground-truth segmentation masks to select object points, but object centers are still estimated from RGB-D point clouds; in the real world, we replace masks with in-the-wild pose estimates from sensor data. The discrete keypose interface also imposes a precision ceiling: \ours{}'s weakest TWIN tasks are Straighten Rope ($34.3\%$), Pick Laptop ($29.0\%$), and Tray Oven ($36.0\%$), which demand fine continuous contact, thin-object grasping, or constrained multi-step manipulation. A finer $200^3$ voxel grid with $2.5^\circ$ rotation bins (\Cref{sec:ablation_resolution}) only partially closes the gap, indicating that the discretized action representation can be too coarse for delicate manipulation regardless of data. Leader-follower conditioning also brings limited benefit when the arms act almost independently, as in Item Drawer and Bottle Fridge, where \ours{} does not exceed the strongest SA/DA baseline. Finally, the approach is bounded by the LLM context window and API latency, which limit the number of demonstrations and the horizon of feasible tasks (\cref{sec:supp_latency} quantifies this cost). The sequential two-arm chain may also not scale to larger teams, where later agents could become overly constrained by earlier predictions. Inspired by multi-agent planners such as RoCo~\cite{mandi2024roco}, a promising extension replaces the single chain with a task-dependent coordination graph in which agents exchange tentative subgoals, condition only on tightly coupled partners, and commit after a lightweight consistency check.

\section{Conclusion}
\label{sec:conclusion}

We presented \ours{}, the first in-context learning framework for bimanual robotic manipulation. The core contribution is a leader-follower decomposition that factors bimanual action prediction into two sequential single-arm predictions linked by explicit conditioning: the follower arm observes the leader's planned trajectory as part of its input and synchronizes accordingly. This decomposition addresses the dual challenge of maintaining low prediction dimensionality while preserving inter-arm coordination, enabling off-the-shelf LLMs to serve as effective bimanual policies without any fine-tuning. 
Evaluated on 13 tasks from the TWIN benchmark, BiCICLe achieves the strongest average performance among training-free baselines and surpasses several supervised methods. Furthermore, we demonstrated its strong out-of-distribution generalization on novel tasks, and showed that the same pose-based ICL interface transfers directly from simulation to a physical bimanual Franka Panda system: without any hardware-specific retraining, BiCICLe achieves 53.3\% average success across three real-world tasks, outperforming all real-world ICL baselines across tasks spanning the full range of bimanual coupling requirements.

\acknowledgments{This work was supported by the Sapienza grant RG123188B3EF6A80 (CENTS), the German Research Foundation Emmy Noether Programme (CH 2676/1-1), the EU’s Horizon Europe project ``ARISE" (Grant 101135959), the German Federal Ministry of Research, Technology and Space of Germany project ``RIG” (Grant 16ME1001) and the European Research Council project “SIREN” (Grant 101163933). We also thank CINECA for the allocation of computational resources.}

\bibliography{main}

@String(CVPR  = {IEEE Conf. Comput. Vis. Pattern Recog.})

@String(ICCV  = {Int. Conf. Comput. Vis.})

@String(ICLR  = {Int. Conf. Learn. Represent.})

@String(CVPR  = {CVPR})

@String(ICCV  = {ICCV})

@String(ICLR  = {ICLR})

@String(ICRA  = {ICRA})

@String(IROS  = {IROS})

@INPROCEEDINGS{roboprompt,
  author={Yin, Yida and Wang, Zekai and Sharma, Yuvan and Niu, Dantong and Darrell, Trevor and Herzig, Roei},
  booktitle={2025 IEEE International Conference on Robotics and Automation (ICRA)}, 
  title={In-Context Learning Enables Robot Action Prediction in {LLMs}}, 
  year={2025},
  volume={},
  number={},
  pages={8972-8979},
  doi={10.1109/ICRA55743.2025.11128807}}

@inproceedings{kat,
title={Keypoint Action Tokens Enable In-Context Imitation Learning in Robotics},
author={Di Palo, Norman and Johns, Edward},
booktitle = {Proceedings of Robotics: Science and Systems (RSS)},
year={2024},
}

@inproceedings{chen2016xgboost,
author = {Chen, Tianqi and Guestrin, Carlos},
title = {XGBoost: A Scalable Tree Boosting System},
year = {2016},
isbn = {9781450342322},
publisher = {Association for Computing Machinery},
address = {New York, NY, USA},
url = {https://doi.org/10.1145/2939672.2939785},
doi = {10.1145/2939672.2939785},
booktitle = {Proceedings of the 22nd ACM SIGKDD International Conference on Knowledge Discovery and Data Mining},
pages = {785-794},
numpages = {10},
location = {San Francisco, California, USA},
series = {KDD '16}
}

@inproceedings{brown2020language,
 author = {Brown, Tom and Mann, Benjamin and Ryder, Nick and Subbiah, Melanie and Kaplan, Jared D and Dhariwal, Prafulla and Neelakantan, Arvind and Shyam, Pranav and Sastry, Girish and Askell, Amanda and Agarwal, Sandhini and Herbert-Voss, Ariel and Krueger, Gretchen and Henighan, Tom and Child, Rewon and Ramesh, Aditya and Ziegler, Daniel and Wu, Jeffrey and Winter, Clemens and Hesse, Chris and Chen, Mark and Sigler, Eric and Litwin, Mateusz and Gray, Scott and Chess, Benjamin and Clark, Jack and Berner, Christopher and McCandlish, Sam and Radford, Alec and Sutskever, Ilya and Amodei, Dario},
 booktitle = {Advances in Neural Information Processing Systems},
 editor = {H. Larochelle and M. Ranzato and R. Hadsell and M.F. Balcan and H. Lin},
 pages = {1877--1901},
 publisher = {Curran Associates, Inc.},
 title = {Language Models are Few-Shot Learners},
 url = {https://proceedings.neurips.cc/paper_files/paper/2020/file/1457c0d6bfcb4967418bfb8ac142f64a-Paper.pdf},
 volume = {33},
 year = {2020}
}

@misc{singh2025openaigpt5card,
      title={{OpenAI} {GPT-5} System Card}, 
      author={Aaditya Singh and others},
      year={2025},
      eprint={2601.03267},
      archivePrefix={arXiv},
      primaryClass={cs.CL},
      url={https://arxiv.org/abs/2601.03267}, 
}

@inproceedings{wei2022chain,
author = {Wei, Jason and Wang, Xuezhi and Schuurmans, Dale and Bosma, Maarten and Ichter, Brian and Xia, Fei and Chi, Ed H. and Le, Quoc V. and Zhou, Denny},
title = {Chain-of-thought prompting elicits reasoning in large language models},
year = {2022},
isbn = {9781713871088},
publisher = {Curran Associates Inc.},
address = {Red Hook, NY, USA},
booktitle = {Proceedings of the 36th International Conference on Neural Information Processing Systems},
articleno = {1800},
numpages = {14},
location = {New Orleans, LA, USA},
series = {NIPS '22}
}

@inproceedings{
wang2023selfconsistency,
title={Self-Consistency Improves Chain of Thought Reasoning in Language Models},
author={Xuezhi Wang and Jason Wei and Dale Schuurmans and Quoc V Le and Ed H. Chi and Sharan Narang and Aakanksha Chowdhery and Denny Zhou},
booktitle={The Eleventh International Conference on Learning Representations },
year={2023},
url={https://openreview.net/forum?id=1PL1NIMMrw}
}

@inproceedings{saycan2022arxiv,
  title={Do As {I} Can, Not As {I} Say: Grounding Language in Robotic Affordances},
  author={Michael Ahn and Anthony Brohan and Noah Brown and Yevgen Chebotar and Omar Cortes and Byron David and Chelsea Finn and Keerthana Gopalakrishnan and Karol Hausman and Alexander Herzog and Daniel Ho and Jasmine Hsu and Julian Ibarz and Brian Ichter and Alex Irpan and Eric Jang and Rosario M Jauregui Ruano and Kyle Jeffrey and Sally Jesmonth and Nikhil Jayant Joshi and Ryan C. Julian and Dmitry Kalashnikov and Yuheng Kuang and Kuang-Huei Lee and Sergey Levine and Yao Lu and Linda Luu and Carolina Parada and Peter Pastor and Jornell Quiambao and Kanishka Rao and Jarek Rettinghouse and Diego M Reyes and Pierre Sermanet and Nicolas Sievers and Clayton Tan and Alexander Toshev and Vincent Vanhoucke and F. Xia and Ted Xiao and Peng Xu and Sichun Xu and Mengyuan Yan},
  booktitle={Conference on Robot Learning},
  year={2022},
  url={https://api.semanticscholar.org/CorpusID:247939706}
}

@InProceedings{huang2022language,
  title = 	 {Language Models as Zero-Shot Planners: Extracting Actionable Knowledge for Embodied Agents},
  author =       {Huang, Wenlong and Abbeel, Pieter and Pathak, Deepak and Mordatch, Igor},
  booktitle = 	 {Proceedings of the 39th International Conference on Machine Learning},
  pages = 	 {9118--9147},
  year = 	 {2022},
  editor = 	 {Chaudhuri, Kamalika and Jegelka, Stefanie and Song, Le and Szepesvari, Csaba and Niu, Gang and Sabato, Sivan},
  volume = 	 {162},
  series = 	 {Proceedings of Machine Learning Research},
  month = 	 {17--23 Jul},
  publisher =    {PMLR},
  url = 	 {https://proceedings.mlr.press/v162/huang22a.html}
}

@INPROCEEDINGS{liang2023code,
  author={Liang, Jacky and Huang, Wenlong and Xia, Fei and Xu, Peng and Hausman, Karol and Ichter, Brian and Florence, Pete and Zeng, Andy},
  booktitle={2023 IEEE International Conference on Robotics and Automation (ICRA)}, 
  title={Code as Policies: Language Model Programs for Embodied Control}, 
  year={2023},
  volume={},
  number={},
  pages={9493-9500},
  doi={10.1109/ICRA48891.2023.10160591}}

@article{fu2026capx,
  title     = {{CaP-X}: A Framework for Benchmarking and Improving Coding Agents for Robot Manipulation},
  author    = {Fu, Max and Yu, Justin and El-Refai, Karim and Kou, Ethan and Xue, Haoru and Huang, Huang and Xiao, Wenli and Wang, Guanzhi and Li, Fei-Fei and Shi, Guanya and Wu, Jiajun and Sastry, Shankar and Zhu, Yuke and Goldberg, Ken and Fan, Jim},
  journal   = {arXiv preprint arXiv:2603.22435},
  year      = {2026},
  url       = {https://arxiv.org/abs/2603.22435}
}

@INPROCEEDINGS{singh2023progprompt,
  author={Singh, Ishika and Blukis, Valts and Mousavian, Arsalan and Goyal, Ankit and Xu, Danfei and Tremblay, Jonathan and Fox, Dieter and Thomason, Jesse and Garg, Animesh},
  booktitle={2023 IEEE International Conference on Robotics and Automation (ICRA)}, 
  title={{ProgPrompt}: Generating Situated Robot Task Plans using Large Language Models}, 
  year={2023},
  volume={},
  number={},
  pages={11523-11530},
  doi={10.1109/ICRA48891.2023.10161317}}

@ARTICLE{vemprala2024chatgpt,
  title={{ChatGPT} for Robotics: Design Principles and Model Abilities},
  author={Sai H. Vemprala and Rogerio Bonatti and Arthur Fender C. Bucker and Ashish Kapoor},
  journal={IEEE Access},
  year={2023},
  volume={12},
  pages={55682-55696},
  url={https://api.semanticscholar.org/CorpusID:259141622}
}

@InProceedings{brohan2023rt2,
  title = 	 {{RT-2}: Vision-Language-Action Models Transfer Web Knowledge to Robotic Control},
  author =       {Zitkovich, Brianna and Yu, Tianhe and Xu, Sichun and Xu, Peng and Xiao, Ted and Xia, Fei and Wu, Jialin and Wohlhart, Paul and Welker, Stefan and Wahid, Ayzaan and Vuong, Quan and Vanhoucke, Vincent and Tran, Huong and Soricut, Radu and Singh, Anikait and Singh, Jaspiar and Sermanet, Pierre and Sanketi, Pannag R. and Salazar, Grecia and Ryoo, Michael S. and Reymann, Krista and Rao, Kanishka and Pertsch, Karl and Mordatch, Igor and Michalewski, Henryk and Lu, Yao and Levine, Sergey and Lee, Lisa and Lee, Tsang-Wei Edward and Leal, Isabel and Kuang, Yuheng and Kalashnikov, Dmitry and Julian, Ryan and Joshi, Nikhil J. and Irpan, Alex and Ichter, Brian and Hsu, Jasmine and Herzog, Alexander and Hausman, Karol and Gopalakrishnan, Keerthana and Fu, Chuyuan and Florence, Pete and Finn, Chelsea and Dubey, Kumar Avinava and Driess, Danny and Ding, Tianli and Choromanski, Krzysztof Marcin and Chen, Xi and Chebotar, Yevgen and Carbajal, Justice and Brown, Noah and Brohan, Anthony and Arenas, Montserrat Gonzalez and Han, Kehang},
  booktitle = 	 {Proceedings of The 7th Conference on Robot Learning},
  pages = 	 {2165--2183},
  year = 	 {2023},
  editor = 	 {Tan, Jie and Toussaint, Marc and Darvish, Kourosh},
  volume = 	 {229},
  series = 	 {Proceedings of Machine Learning Research},
  month = 	 {06--09 Nov},
  publisher =    {PMLR},
  url = 	 {https://proceedings.mlr.press/v229/zitkovich23a.html}
}

@InProceedings{kim2024openvla,
  title = 	 {{OpenVLA}: An Open-Source Vision-Language-Action Model},
  author =       {Kim, Moo Jin and Pertsch, Karl and Karamcheti, Siddharth and Xiao, Ted and Balakrishna, Ashwin and Nair, Suraj and Rafailov, Rafael and Foster, Ethan P and Sanketi, Pannag R and Vuong, Quan and Kollar, Thomas and Burchfiel, Benjamin and Tedrake, Russ and Sadigh, Dorsa and Levine, Sergey and Liang, Percy and Finn, Chelsea},
  booktitle = 	 {Proceedings of The 8th Conference on Robot Learning},
  pages = 	 {2679--2713},
  year = 	 {2025},
  editor = 	 {Agrawal, Pulkit and Kroemer, Oliver and Burgard, Wolfram},
  volume = 	 {270},
  series = 	 {Proceedings of Machine Learning Research},
  month = 	 {06--09 Nov},
  publisher =    {PMLR},
  url = 	 {https://proceedings.mlr.press/v270/kim25c.html},
}

@misc{li2024robovlms,
      title={What Matters in Building Vision-Language-Action Models for Generalist Robots}, 
      author={Xinghang Li and Peiyan Li and Long Qian and Minghuan Liu and Dong Wang and Jirong Liu and Bingyi Kang and Xiao Ma and Xinlong Wang and Di Guo and Tao Kong and Hanbo Zhang and Huaping Liu},
      year={2026},
      eprint={2412.14058},
      archivePrefix={arXiv},
      primaryClass={cs.RO},
      url={https://arxiv.org/abs/2412.14058}, 
}

@INPROCEEDINGS{lee2015learning,
  author={Lee, Alex X. and Lu, Henry and Gupta, Abhishek and Levine, Sergey and Abbeel, Pieter},
  booktitle={2015 IEEE International Conference on Robotics and Automation (ICRA)}, 
  title={Learning force-based manipulation of deformable objects from multiple demonstrations}, 
  year={2015},
  volume={},
  number={},
  pages={177-184},
  doi={10.1109/ICRA.2015.7138997}}

@INPROCEEDINGS{chitnis2020efficient,
  author={Chitnis, Rohan and Tulsiani, Shubham and Gupta, Saurabh and Gupta, Abhinav},
  booktitle={2020 IEEE International Conference on Robotics and Automation (ICRA)}, 
  title={Efficient Bimanual Manipulation Using Learned Task Schemas}, 
  year={2020},
  volume={},
  number={},
  pages={1149-1155},
  doi={10.1109/ICRA40945.2020.9196958}}

@inproceedings{xie2020deep,
author = {Xie, Fan and Chowdhury, Alexander and De Paolis Kaluza, M. Clara and Zhao, Linfeng and Wong, Lawson L.S. and Yu, Rose},
title = {Deep imitation learning for bimanual robotic manipulation},
year = {2020},
isbn = {9781713829546},
publisher = {Curran Associates Inc.},
address = {Red Hook, NY, USA},
booktitle = {Proceedings of the 34th International Conference on Neural Information Processing Systems},
articleno = {196},
numpages = {11},
location = {Vancouver, BC, Canada},
series = {NIPS '20}
}

@InProceedings{grannen2023stabilize,
  title = 	 {Stabilize to Act: Learning to Coordinate for Bimanual Manipulation},
  author =       {Grannen, Jennifer and Wu, Yilin and Vu, Brandon and Sadigh, Dorsa},
  booktitle = 	 {Proceedings of The 7th Conference on Robot Learning},
  pages = 	 {563--576},
  year = 	 {2023},
  editor = 	 {Tan, Jie and Toussaint, Marc and Darvish, Kourosh},
  volume = 	 {229},
  series = 	 {Proceedings of Machine Learning Research},
  month = 	 {06--09 Nov},
  publisher =    {PMLR},
  url = 	 {https://proceedings.mlr.press/v229/grannen23a.html},
}

@INPROCEEDINGS{lee2024bimact,
  author={Buamanee, Thanpimon and Kobayashi, Masato and Uranishi, Yuki and Takemura, Haruo},
  booktitle={2024 IEEE International Conference on Advanced Intelligent Mechatronics (AIM)}, 
  title={{Bi-ACT}: Bilateral Control-Based Imitation Learning via Action Chunking with Transformer}, 
  year={2024},
  volume={},
  number={},
  pages={410-415},
  doi={10.1109/AIM55361.2024.10637173}}

@inproceedings{chen2022towards,
 author = {Chen, Yuanpei and Wu, Tianhao and Wang, Shengjie and Feng, Xidong and Jiang, Jiechuan and Lu, Zongqing and McAleer, Stephen and Dong, Hao and Zhu, Song-Chun and Yang, Yaodong},
 booktitle = {Advances in Neural Information Processing Systems},
 editor = {S. Koyejo and S. Mohamed and A. Agarwal and D. Belgrave and K. Cho and A. Oh},
 pages = {5150--5163},
 publisher = {Curran Associates, Inc.},
 title = {Towards Human-Level Bimanual Dexterous Manipulation with Reinforcement Learning},
 url = {https://proceedings.neurips.cc/paper_files/paper/2022/file/217a2a387f52c30755c37b0a73430291-Paper-Datasets_and_Benchmarks.pdf},
 volume = {35},
 year = {2022}
}

@inproceedings{
im2026twinvla,
title={Twin{VLA}: Data-Efficient Bimanual Manipulation with Twin Single-Arm Vision-Language-Action Models},
author={Hokyun Im and Euijin Jeong and Andrey Kolobov and Jianlong Fu and Youngwoon Lee},
booktitle={The Fourteenth International Conference on Learning Representations},
year={2026},
url={https://openreview.net/forum?id=jG9W6nAwVz}
}

@inproceedings{zhao2023act,
  title={Learning Fine-Grained Bimanual Manipulation with Low-Cost Hardware},
  author={Tony Z. Zhao and Vikash Kumar and Sergey Levine and Chelsea Finn},
  booktitle={Proceedings of Robotics: Science and Systems (RSS)},
  year={2023}
}

@article{smith2012dual,
title = {Dual arm manipulation—A survey},
journal = {Robotics and Autonomous Systems},
volume = {60},
number = {10},
pages = {1340-1353},
year = {2012},
issn = {0921-8890},
doi = {https://doi.org/10.1016/j.robot.2012.07.005},
url = {https://www.sciencedirect.com/science/article/pii/S092188901200108X},
author = {Christian Smith and Yiannis Karayiannidis and Lazaros Nalpantidis and Xavi Gratal and Peng Qi and Dimos V. Dimarogonas and Danica Kragic},
}

@INPROCEEDINGS{koga1994multi,
  author={Koga, Y. and Latombe, J.-C.},
  booktitle={Proceedings of the 1994 IEEE International Conference on Robotics and Automation}, 
  title={On multi-arm manipulation planning}, 
  year={1994},
  volume={},
  number={},
  pages={945-952 vol.2},
  doi={10.1109/ROBOT.1994.351231}}

@article{amir2021deep,
	  author    = {Shir Amir and Yossi Gandelsman and Shai Bagon and Tali Dekel},
  	  title     = {Deep {ViT} Features as Dense Visual Descriptors}, 
	  journal   = {ECCVW What is Motion For?},
	  year      = {2022},
  }

@InProceedings{caron2021emerging,
    author    = {Caron, Mathilde and Touvron, Hugo and Misra, Ishan and J\'egou, Herv\'e and Mairal, Julien and Bojanowski, Piotr and Joulin, Armand},
    title     = {Emerging Properties in Self-Supervised Vision Transformers},
    booktitle = {Proceedings of the IEEE/CVF International Conference on Computer Vision (ICCV)},
    month     = {October},
    year      = {2021},
    pages     = {9650-9660}
}

@inproceedings{alayrac2022flamingo,
author = {Alayrac, Jean-Baptiste and Donahue, Jeff and Luc, Pauline and Miech, Antoine and Barr, Iain and Hasson, Yana and Lenc, Karel and Mensch, Arthur and Millicah, Katie and Reynolds, Malcolm and Ring, Roman and Rutherford, Eliza and Cabi, Serkan and Han, Tengda and Gong, Zhitao and Samangooei, Sina and Monteiro, Marianne and Menick, Jacob and Borgeaud, Sebastian and Brock, Andrew and Nematzadeh, Aida and Sharifzadeh, Sahand and Binkowski, Mikolaj and Barreira, Ricardo and Vinyals, Oriol and Zisserman, Andrew and Simonyan, Karen},
title = {Flamingo: a visual language model for few-shot learning},
year = {2022},
isbn = {9781713871088},
publisher = {Curran Associates Inc.},
address = {Red Hook, NY, USA},
booktitle = {Proceedings of the 36th International Conference on Neural Information Processing Systems},
articleno = {1723},
numpages = {21},
location = {New Orleans, LA, USA},
series = {NIPS '22}
}

@inproceedings{
zhang2023building,
title={Building Cooperative Embodied Agents Modularly with Large Language Models},
author={Hongxin Zhang and Weihua Du and Jiaming Shan and Qinhong Zhou and Yilun Du and Joshua B. Tenenbaum and Tianmin Shu and Chuang Gan},
booktitle={The Twelfth International Conference on Learning Representations},
year={2024},
url={https://openreview.net/forum?id=EnXJfQqy0K}
}

@inproceedings{mandi2024roco,
  author       = {Zhao Mandi and
                  Shreeya Jain and
                  Shuran Song},
  title        = {{RoCo}: Dialectic Multi-Robot Collaboration with Large Language Models},
  booktitle    = {{IEEE} International Conference on Robotics and Automation, {ICRA}
                  2024, Yokohama, Japan, May 13-17, 2024},
  pages        = {286--299},
  publisher    = {{IEEE}},
  year         = {2024},
  url          = {https://doi.org/10.1109/ICRA57147.2024.10610855},
  doi          = {10.1109/ICRA57147.2024.10610855},
  bibsource    = {dblp computer science bibliography, https://dblp.org}
}

@article{james2020rlbench,
  title={{RLBench}: The Robot Learning Benchmark \& Learning Environment},
  author={Stephen James and Zicong Ma and David Rovick Arrojo and Andrew J. Davison},
  journal={IEEE Robotics and Automation Letters},
  year={2019},
  volume={5},
  pages={3019-3026},
  url={https://api.semanticscholar.org/CorpusID:202889132}
}

@article{twin,
  title={{TWIN}: Two-handed Intelligent Benchmark for Bimanual Manipulation},
  author={Markus Grotz and Mohit Shridhar and Yu-Wei Chao and Tamim Asfour and Dieter Fox},
  journal={2025 IEEE International Conference on Robotics and Automation (ICRA)},
  year={2025},
  pages={7945-7951},
  url={https://api.semanticscholar.org/CorpusID:281094284}
}

@article{open3d,
    author    = {Qian-Yi Zhou and Jaesik Park and Vladlen Koltun},
    title     = {{Open3D}: {A} Modern Library for {3D} Data Processing},
    journal   = {arXiv:1801.09847},
    year      = {2018},
}

@article{ze20253dfa,
            author = {Gkanatsios, Nikolaos and Xu, Jiahe and Bronars, Matthew and Mousavian, Arsalan and Ke, Tsung-Wei and Fragkiadaki, Katerina},
            title = {{3D FlowMatch Actor}: Unified {3D} Policy for Single- and Dual-Arm Manipulation},
            journal = {Arxiv},
            year = {2025}
        }

@InProceedings{shridhar2022peract,
  title = 	 {{Perceiver-Actor}: A Multi-Task Transformer for Robotic Manipulation},
  author =       {Shridhar, Mohit and Manuelli, Lucas and Fox, Dieter},
  booktitle = 	 {Proceedings of The 6th Conference on Robot Learning},
  pages = 	 {785--799},
  year = 	 {2023},
  editor = 	 {Liu, Karen and Kulic, Dana and Ichnowski, Jeff},
  volume = 	 {205},
  series = 	 {Proceedings of Machine Learning Research},
  month = 	 {14--18 Dec},
  publisher =    {PMLR},
  url = 	 {https://proceedings.mlr.press/v205/shridhar23a.html},
}

@inproceedings{ze2024dp3,
    	title={{3D Diffusion} Policy: Generalizable Visuomotor Policy Learning via Simple {3D} Representations},
	author={Yanjie Ze and Gu Zhang and Kangning Zhang and Chenyuan Hu and Muhan Wang and Huazhe Xu},
	booktitle={Proceedings of Robotics: Science and Systems (RSS)},
	year={2024}
}

@inproceedings{kstardiffuser2024,
  title={Spatial-Temporal Graph Diffusion Policy with Kinematic Modeling for Bimanual Robotic Manipulation},
  author={Lv, Qi and Li, Hao and Deng, Xiang and Shao, Rui and Li, Yinchuan and Hao, Jianye and Gao, Longxiang and Wang, Michael Yu and Nie, Liqiang},
  booktitle={Proceedings of the IEEE/CVF Conference on Computer Vision and Pattern Recognition (CVPR)},
  year={2025},
  pages={17394-17404},
  doi={10.1109/CVPR52734.2025.01621}
}

@inproceedings{ppi2024,
  title={Gripper Keypose and Object Pointflow as Interfaces for Bimanual Robotic Manipulation},
  author={Yang, Yuyin and Cai, Zetao and Tian, Yang and Zeng, Jia and Pang, Jiangmiao},
  booktitle={Proceedings of Robotics: Science and Systems (RSS)},
  year={2025}
}

@article{anybimanual2024,
  title={{AnyBimanual}: Transferring Unimanual Policy for General Bimanual Manipulation},
  author={Lu, Guanxing and Yu, Tengbo and Deng, Haoyuan and Chen, Season Si and Tang, Yansong and Wang, Ziwei},
  journal={arXiv preprint arXiv:2412.06779},
  year={2024}
}

@ARTICLE{krebs2022tax,
  author={Krebs, Franziska and Asfour, Tamim},
  journal={IEEE Robotics and Automation Letters}, 
  title={A Bimanual Manipulation Taxonomy}, 
  year={2022},
  volume={7},
  number={4},
  pages={11031-11038},
  doi={10.1109/LRA.2022.3196158}
}

@inproceedings{zhang2026vlm4vla,
  title={{VLM4VLA}: Revisiting Vision-Language-Models in Vision-Language-Action Models},
  author={Zhang, Jianke and Chen, Xiaoyu and Wang, Qiuyue and Li, Mingsheng and Guo, Yanjiang and Hu, Yucheng and Zhang, Jiajun and Bai, Shuai and Lin, Junyang},
  booktitle={International Conference on Learning Representations (ICLR)},
  year={2026},
}

@inproceedings{coppeliasim,
    author="E. Rohmer and S. P. N. Singh and M. Freese",
    title="{CoppeliaSim} (formerly {V-REP}): a Versatile and
        Scalable Robot Simulation Framework",
    booktitle="Proc. of The International Conference on
        Intelligent Robots and Systems (IROS)",
    year="2013",
    note="www.coppeliarobotics.com"
}

@inproceedings{
sridhar2025ricl,
title={{RICL}:  Adding In-Context Adaptability to Pre-Trained Vision-Language-Action Models},
author={Kaustubh Sridhar and Souradeep Dutta and Dinesh Jayaraman and Insup Lee},
booktitle={9th Annual Conference on Robot Learning},
year={2025},
url={https://openreview.net/forum?id=6AASPlloSt}
}

@misc{qwen2025qwen25technicalreport,
      title={Qwen2.5 Technical Report}, 
      author={An Yang and Baosong Yang and Beichen Zhang and Binyuan Hui and Bo Zheng and Bowen Yu and Chengyuan Li and Dayiheng Liu and Fei Huang and Haoran Wei and Huan Lin and Jian Yang and Jianhong Tu and Jianwei Zhang and Jianxin Yang and Jiaxi Yang and Jingren Zhou and Junyang Lin and Kai Dang and Keming Lu and Keqin Bao and Kexin Yang and Le Yu and Mei Li and Mingfeng Xue and Pei Zhang and Qin Zhu and Rui Men and Runji Lin and Tianhao Li and Tianyi Tang and Tingyu Xia and Xingzhang Ren and Xuancheng Ren and Yang Fan and Yang Su and Yichang Zhang and Yu Wan and Yuqiong Liu and Zeyu Cui and Zhenru Zhang and Zihan Qiu},
      year={2025},
      eprint={2412.15115},
      archivePrefix={arXiv},
      primaryClass={cs.CL},
      url={https://arxiv.org/abs/2412.15115}, 
}

@misc{bai2025qwen25vltechnicalreport,
      title={{Qwen2.5-VL} Technical Report}, 
      author={Shuai Bai and Keqin Chen and Xuejing Liu and Jialin Wang and Wenbin Ge and Sibo Song and Kai Dang and Peng Wang and Shijie Wang and Jun Tang and Humen Zhong and Yuanzhi Zhu and Mingkun Yang and Zhaohai Li and Jianqiang Wan and Pengfei Wang and Wei Ding and Zheren Fu and Yiheng Xu and Jiabo Ye and Xi Zhang and Tianbao Xie and Zesen Cheng and Hang Zhang and Zhibo Yang and Haiyang Xu and Junyang Lin},
      year={2025},
      eprint={2502.13923},
      archivePrefix={arXiv},
      primaryClass={cs.CV},
      url={https://arxiv.org/abs/2502.13923}, 
}

@article{pertsch2025fast,
  title={{FAST}: Efficient action tokenization for vision-language-action models},
  author={Pertsch, Karl and Stachowicz, Kyle and Ichter, Brian and Driess, Danny and Nair, Suraj and Vuong, Quan and Mees, Oier and Finn, Chelsea and Levine, Sergey},
  journal={arXiv preprint arXiv:2501.09747},
  year={2025}
}

@inproceedings{droid,
title = "{DROID}: A large-scale in-the-wild robot manipulation dataset",
author = "Alexander Khazatsky and others",
year = "2024",
month = may,
day = "14",
language = "English",
series = "Robotics: science and systems",
booktitle = "Proceedings of Robotics: Science and Systems XX",
note = "Robotics: Science and Systems, R:SS ; Conference date: 15-07-2024 Through 19-07-2024",
url = "https://roboticsconference.org/",
}

@INPROCEEDINGS{BlackK-RSS-25,
    AUTHOR    = {Kevin Black AND Noah Brown AND Danny Driess AND Adnan Esmail AND Michael Robert Equi AND Chelsea Finn AND Niccolo Fusai AND Lachy Groom AND Karol Hausman AND Brian Ichter AND Szymon Jakubczak AND Tim Jones AND Liyiming Ke AND Sergey Levine AND Adrian Li-Bell AND Mohith Mothukuri AND Suraj Nair AND Karl Pertsch AND Lucy Xiaoyang Shi AND Laura Smith AND James Tanner AND Quan Vuong AND Anna Walling AND Haohuan Wang AND Ury Zhilinsky},
    TITLE     = {{$\pi_0$: A Vision-Language-Action Flow Model for General Robot Control}},
    BOOKTITLE = {Proceedings of Robotics: Science and Systems},
    YEAR      = {2025},
    ADDRESS   = {Los Angeles, CA, USA},
    MONTH     = {June},
    DOI       = {10.15607/RSS.2025.XXI.010}
}

@article{Wen2023FoundationPoseU6,
  title={{FoundationPose}: Unified {6D} Pose Estimation and Tracking of Novel Objects},
  author={Bowen Wen and Wei Yang and Jan Kautz and Stanley T. Birchfield},
  journal={2024 IEEE/CVF Conference on Computer Vision and Pattern Recognition (CVPR)},
  year={2023},
  pages={17868-17879},
  url={https://api.semanticscholar.org/CorpusID:266191252}
}

@article{coleman2014reducing,
  title={Reducing the Barrier to Entry of Complex Robotic Software: a {MoveIt!} Case Study},
  author={Coleman, David T and Sucan, Ioan A and Chitta, Sachin and Correll, Nikolaus},
  journal={JOURNAL OF SOFTWARE ENGINEERING IN ROBOTICS},
  volume={5},
  number={1},
  pages={3--16},
  year={2014},
  publisher={Universit{\`a} degli studi di Bergamo}
}

@article{shah2025mimicdroid,
  title={{Mimicdroid}: In-context learning for humanoid robot manipulation from human play videos},
  author={Shah, Rutav and Liu, Shuijing and Wang, Qi and Jiang, Zhenyu and Kumar, Sateesh and Seo, Mingyo and Mart{\'\i}n-Mart{\'\i}n, Roberto and Zhu, Yuke},
  journal={arXiv preprint arXiv:2509.09769},
  year={2025}
}

@inproceedings{hu2022lora,
  title={Lo{RA}: Low-Rank Adaptation of Large Language Models},
  author={Edward J. Hu and Yelong Shen and Phillip Wallis and Zeyuan Allen-Zhu and Yuanzhi Li and Shean Wang and Lu Wang and Weizhu Chen},
  booktitle={International Conference on Learning Representations},
  year={2022},
  url={https://openreview.net/forum?id=nZeVKeeFYf9}
}

@InProceedings{pmlr-v229-mirchandani23a,
  title = {Large Language Models as General Pattern Machines},
  author = {Mirchandani, Suvir and Xia, Fei and Florence, Pete and Ichter, Brian and Driess, Danny and Arenas, Montserrat Gonzalez and Rao, Kanishka and Sadigh, Dorsa and Zeng, Andy},
  booktitle = {Proceedings of The 7th Conference on Robot Learning},
  pages = {2498--2518},
  year = {2023},
  editor = {Tan, Jie and Toussaint, Marc and Darvish, Kourosh},
  volume = {229},
  series = {Proceedings of Machine Learning Research},
  month = {06--09 Nov},
  publisher = {PMLR},
  url = {https://proceedings.mlr.press/v229/mirchandani23a.html}
}

\clearpage
\appendix
\section*{Appendix}
\noindent We supplement the main paper submission with the following details:
\begin{itemize}[nosep]
    \item \cref{sec:supp_ablations} provides extended ablation studies on backbone agnosticism, leader arm assignment, sequential prediction, test-time scaling, observation representations, discretization resolution, and point-cloud extraction;
    \item \cref{sec:supp_qualitatives} provides additional qualitative simulation analysis;
    \item \cref{sec:supp_ricl} describes the adaptation of an ICL-capable Vision-Language-Action model for bimanual manipulation and compares it with \ours;
    \item \cref{sec:supp_latency} reports LLM call statistics and inference latency;
    \item \cref{sec:supp_prompts} lists the prompt templates used by \ours{}.
\end{itemize}

\section{Ablation Studies}
\label{sec:supp_ablations}

We ablate the key design choices of \ours{} on the TWIN benchmark. Unless stated otherwise, all ablation experiments use GPT-5-mini as the backbone and a single evaluation seed (100 episodes per task).

\subsection{Backbone Agnosticism}
\label{sec:ablation_backbone}

To verify that \ours{} does not rely on the emergent capabilities of a specific proprietary LLM, we replicate our in-context experiments using Qwen~2.5 7B~\cite{qwen2025qwen25technicalreport}, and Qwen~2.5VL 7B~\cite{bai2025qwen25vltechnicalreport} for VLM-LF. Even with this substantially smaller open-source backbone, the relative performance ranking is preserved (\Cref{tab:backbone}). We further leverage this 7B model in \Cref{sec:ablation_leader} to ablate the structural assignment of the leader.

\providecommand{\twintabmeanpad}[1]{\ifdim #1pt<10pt \phantom{00}\else\ifdim #1pt<100pt \phantom{0}\fi\fi}
\providecommand{\twintabstdpad}[1]{\ifdim #1pt<10pt \phantom{0}\fi}
\providecommand{\twintabmean}[1]{\twintabmeanpad{#1}#1}
\providecommand{\twintabstd}[1]{\twintabstdpad{#1}#1}
\providecommand{\twintabnum}[1]{\ensuremath{\twintabmean{#1}}}
\providecommand{\twintabbestnum}[1]{\ensuremath{\twintabmeanpad{#1}\bm{#1}}}
\providecommand{\twintabsecondnum}[1]{\ensuremath{\twintabmeanpad{#1}\underline{#1}}}
\providecommand{\twintabres}[2]{\ensuremath{\twintabmean{#1}_{\pm\twintabstd{#2}}}}
\providecommand{\twintabbestres}[2]{\ensuremath{\twintabmeanpad{#1}\bm{#1}_{\bm{\pm}\twintabstdpad{#2}\bm{#2}}}}
\providecommand{\twintabsecondres}[2]{\ensuremath{\twintabmeanpad{#1}\underline{#1_{\pm\twintabstd{#2}}}}}
\begin{table}[h]
\caption{\textbf{Backbone agnosticism: Qwen~2.5 7B~\cite{qwen2025qwen25technicalreport}.} Success rates (\%) on the TWIN benchmark using a 7B open-source backbone. Mean $\pm$ std over 3 seeds $\times$ 100 episodes. \textbf{Bold}: best per task. \underline{Underline}: second best.}
\label{tab:backbone}
\centering
\resizebox{\textwidth}{!}{%
\begin{tabular}{l|rrrrrrrrrrrrr|r}
\toprule
\textbf{Method} & \rotatebox{70}{Push Box} & \rotatebox{70}{Dual Buttons} & \rotatebox{70}{Bottle Fridge} & \rotatebox{70}{Handover} & \rotatebox{70}{Handover Easy} & \rotatebox{70}{Lift Ball} & \rotatebox{70}{Lift Tray} & \rotatebox{70}{Pick Laptop} & \rotatebox{70}{Pick Plate} & \rotatebox{70}{Straighten Rope} & \rotatebox{70}{Sweep Dustpan} & \rotatebox{70}{Tray Oven} & \rotatebox{70}{Item Drawer} & \textbf{Avg.} \\
\midrule
VLM-LF & \twintabres{51.0}{4.6} & \twintabres{1.3}{2.3} & \twintabres{3.3}{2.3} & \twintabres{2.3}{1.2} & \twintabres{3.7}{0.5} & \twintabres{3.7}{2.1} & \twintabres{2.7}{1.9} & \twintabres{1.7}{1.5} & \twintabres{3.0}{0.8} & \twintabres{0.0}{0.0} & \twintabres{24.3}{3.2} & \twintabres{4.7}{1.9} & \twintabres{4.0}{1.4} & \twintabnum{8.1} \\
KAT-SA~\cite{kat} & \twintabres{53.3}{1.7} & \twintabres{2.3}{1.5} & \twintabres{6.7}{2.3} & \twintabres{0.0}{0.0} & \twintabres{16.3}{1.7} & \twintabres{34.7}{3.7} & \twintabres{11.0}{0.8} & \twintabres{1.3}{0.5} & \twintabres{2.3}{0.6} & \twintabres{0.0}{0.0} & \twintabres{41.3}{0.6} & \twintabres{3.7}{2.1} & \twintabres{9.3}{4.2} & \twintabnum{14.0} \\
KAT-DA~\cite{kat} & \twintabres{41.7}{3.2} & \twintabsecondres{2.7}{1.9} & \twintabres{14.0}{3.3} & \twintabres{1.0}{0.8} & \twintabres{22.7}{7.2} & \twintabres{12.3}{1.9} & \twintabres{9.0}{0.8} & \twintabres{0.7}{0.5} & \twintabres{1.7}{1.2} & \twintabres{0.3}{0.5} & \twintabres{54.0}{0.8} & \twintabres{2.3}{1.2} & \twintabres{3.0}{2.0} & \twintabnum{12.7} \\
RoboPrompt-SA~\cite{roboprompt} & \twintabbestres{94.7}{0.5} & \twintabbestres{100.0}{0.0} & \twintabres{68.0}{5.0} & \twintabres{82.0}{2.4} & \twintabbestres{62.7}{4.7} & \twintabres{55.7}{1.7} & \twintabres{19.7}{2.9} & \twintabres{5.3}{0.5} & \twintabres{20.7}{0.9} & \twintabres{11.3}{2.5} & \twintabsecondres{91.3}{1.2} & \twintabres{16.3}{0.9} & \twintabres{47.0}{2.2} & \twintabnum{51.9} \\
RoboPrompt-DA~\cite{roboprompt} & \twintabsecondres{94.3}{2.1} & \twintabbestres{100.0}{0.0} & \twintabres{78.7}{2.4} & \twintabres{64.7}{1.7} & \twintabres{38.3}{3.7} & \twintabres{47.0}{2.2} & \twintabres{19.7}{3.1} & \twintabres{6.3}{1.7} & \twintabsecondres{37.3}{1.2} & \twintabres{13.3}{0.9} & \twintabbestres{99.3}{0.5} & \twintabsecondres{21.7}{2.1} & \twintabres{40.7}{2.6} & \twintabnum{51.6} \\
\midrule
\ours{} & \twintabres{83.0}{2.4} & \twintabbestres{100.0}{0.0} & \twintabbestres{83.0}{1.4} & \twintabsecondres{83.7}{1.7} & \twintabsecondres{53.0}{1.6} & \twintabsecondres{56.7}{2.5} & \twintabsecondres{22.7}{4.2} & \twintabbestres{12.0}{1.4} & \twintabbestres{40.3}{3.8} & \twintabsecondres{16.7}{2.1} & \twintabbestres{99.3}{0.5} & \twintabres{20.7}{1.7} & \twintabbestres{51.3}{2.5} & \twintabsecondnum{55.6} \\
+ Best-of-N & \twintabres{92.7}{2.9} & \twintabbestres{100.0}{0.0} & \twintabsecondres{79.7}{1.9} & \twintabbestres{84.3}{2.9} & \twintabres{43.7}{5.3} & \twintabbestres{60.0}{2.2} & \twintabbestres{25.7}{1.2} & \twintabsecondres{9.7}{4.2} & \twintabbestres{40.3}{0.9} & \twintabbestres{18.7}{1.7} & \twintabbestres{99.3}{0.9} & \twintabbestres{30.0}{1.0} & \twintabsecondres{49.7}{5.9} & \twintabbestnum{56.5} \\
\bottomrule
\end{tabular}%
}
\end{table}

As expected, all methods incur a performance drop with the smaller backbone; however, the relative ranking remains consistent: \ours{} with Best-of-N ($56.5\%$) outperforms the best baseline (RoboPrompt-SA, $51.9\%$) by 4.6 percentage points, confirming that the leader-follower decomposition provides an architectural advantage independent of model capacity. The per-task breakdown mirrors the patterns observed with GPT-5-mini: \ours{} dominates on asymmetric and fine-grained tasks (Bottle Fridge, Pick Plate, Pick Laptop, Item Drawer), while Best-of-N yields the largest gains on symmetric tasks requiring precise coordination (Lift Ball $60.0\%$ vs.\ $56.7\%$, Tray Oven $30.0\%$ vs.\ $20.7\%$). Notably, even with a 7B open-source model, \ours{} surpasses several supervised methods including PerAct$^2$ ($16.8\%$) and AnyBimanual ($32.0\%$), reinforcing the viability of training-free ICL across different backbone scales.

\subsection{Leader Arm Assignment}
\label{sec:ablation_leader}

\ours{} designates one arm as the leader and the other as the follower. By default, the right arm leads. To verify that this choice does not introduce a systematic bias, we evaluate a variant in which the left arm leads instead.

\begin{table}[h]
\caption{\textbf{Ablation: Leader arm choice.} Success rates (\%) on the TWIN benchmark for two backbones: GPT-5-mini and Qwen~2.5 7B. Right arm results from main paper. \textbf{Bold}: best per task within each backbone.}
\label{tab:ablation_leader}
\centering
\resizebox{\textwidth}{!}{%
\begin{tabular}{l|rrrrrrrrrrrrr|r}
\toprule
\textbf{Leader} & \rotatebox{70}{Push Box} & \rotatebox{70}{Dual Buttons} & \rotatebox{70}{Bottle Fridge} & \rotatebox{70}{Handover} & \rotatebox{70}{Handover Easy} & \rotatebox{70}{Lift Ball} & \rotatebox{70}{Lift Tray} & \rotatebox{70}{Pick Laptop} & \rotatebox{70}{Pick Plate} & \rotatebox{70}{Straighten Rope} & \rotatebox{70}{Sweep Dustpan} & \rotatebox{70}{Tray Oven} & \rotatebox{70}{Item Drawer} & \textbf{Avg.} \\
\midrule
\multicolumn{15}{l}{\textit{GPT-5-mini~\cite{singh2025openaigpt5card}}} \\
Right (default) & \twintabbestnum{99.0} & \twintabbestnum{100.0} & \twintabnum{80.3} & \twintabbestnum{94.3} & \twintabbestnum{68.0} & \twintabbestnum{83.7} & \twintabbestnum{83.0} & \twintabnum{29.0} & \twintabbestnum{65.3} & \twintabnum{34.3} & \twintabnum{97.3} & \twintabbestnum{36.0} & \twintabnum{46.7} & \twintabbestnum{70.5} \\
Left & \twintabnum{94.0} & \twintabbestnum{100.0} & \twintabbestnum{83.0} & \twintabnum{94.0} & \twintabnum{60.0} & \twintabnum{79.0} & \twintabbestnum{83.0} & \twintabbestnum{30.0} & \twintabnum{57.0} & \twintabbestnum{38.0} & \twintabbestnum{99.0} & \twintabbestnum{36.0} & \twintabbestnum{47.0} & \twintabnum{69.2} \\
\midrule
\multicolumn{15}{l}{\textit{Qwen~2.5 7B~\cite{qwen2025qwen25technicalreport}}} \\
Right (default) & \twintabnum{83.0} & \twintabbestnum{100.0} & \twintabbestnum{83.0} & \twintabbestnum{83.7} & \twintabbestnum{53.0} & \twintabbestnum{56.7} & \twintabbestnum{22.7} & \twintabbestnum{12.0} & \twintabbestnum{40.3} & \twintabbestnum{16.7} & \twintabnum{99.3} & \twintabnum{20.7} & \twintabbestnum{51.3} & \twintabbestnum{55.6} \\
Left & \twintabbestnum{94.0} & \twintabbestnum{100.0} & \twintabnum{81.0} & \twintabnum{82.0} & \twintabnum{52.0} & \twintabnum{51.0} & \twintabnum{18.0} & \twintabnum{5.0} & \twintabnum{35.0} & \twintabnum{15.0} & \twintabbestnum{100.0} & \twintabbestnum{21.0} & \twintabnum{41.0} & \twintabnum{53.5} \\
\bottomrule
\end{tabular}%
}
\end{table}

\Cref{tab:ablation_leader} shows that, for both backbones, switching the leader arm has a modest effect on overall performance. With GPT-5-mini, the gap is only $1.3$ percentage points ($70.5\%$ right versus $69.2\%$ left), and per-task differences are inconsistent in direction: the left arm even outperforms the right on Bottle Fridge ($83.0$ vs $80.3$), Straighten Rope ($38.0$ vs $34.3$), and Sweep Dustpan ($99.0$ vs $97.3$). With Qwen~2.5 7B the gap widens slightly to $2.1$ pp ($55.6\%$ versus $53.5\%$) and the right leader wins on 9 of 13 tasks, yet most individual differences remain within $5.0$ pp; the only large swings are Push Box, where the left leader improves by $+11.0$ pp, and Item Drawer, where the right leader gains $+10.3$ pp.

Across both models, the tasks where the left arm falls most (Handover Easy and Pick Plate for GPT-5-mini; Pick Laptop and Item Drawer for Qwen) are inherently ``right-handed'': the right arm performs the primary manipulation (grasping, lifting, or placing), so designating it as leader naturally aligns the decomposition with the task's role structure. When the left arm leads instead, the follower must execute the more demanding action conditioned on a less informative leader plan. These consistent patterns confirm that the sequential conditioning is largely agnostic to leader assignment across model scales, with residual gaps attributable to the intrinsic handedness of individual tasks rather than an architectural bias.

\subsection{Sequential Arm Prediction}
\label{sec:ablation_sequential}

We evaluate \emph{Sequential Arms} to isolate whether \ours{} gains mainly from reducing the prediction dimensionality or from the leader-follower conditioning itself. Sequential Arms predicts one arm first and the other second, using the same per-arm prompts as RoboPrompt-DA but just one shared agent. Thus, it preserves the sequential $\mathbb{Z}^{7}{+}\mathbb{Z}^{7}$ output structure, but removes the core leader-follower mechanism. RoboPrompt-SA and RoboPrompt-DA provide the two endpoints of this comparison: monolithic $\mathbb{Z}^{14}$ prediction and independent $\mathbb{Z}^{7}{+}\mathbb{Z}^{7}$ prediction.

\begin{table}[h]
\caption{\textbf{Ablation: dimensionality reduction vs.\ leader-follower conditioning.} Success rates (\%) on the TWIN benchmark. RoboPrompt and \ours{} rows from \Cref{tab:main_results}.}
\label{tab:ablation_sequential}
\centering
\resizebox{\textwidth}{!}{%
\begin{tabular}{l|rrrrrrrrrrrrr|r}
\toprule
\textbf{Variant} & \rotatebox{70}{Push Box} & \rotatebox{70}{Dual Buttons} & \rotatebox{70}{Bottle Fridge} & \rotatebox{70}{Handover} & \rotatebox{70}{Handover Easy} & \rotatebox{70}{Lift Ball} & \rotatebox{70}{Lift Tray} & \rotatebox{70}{Pick Laptop} & \rotatebox{70}{Pick Plate} & \rotatebox{70}{Straighten Rope} & \rotatebox{70}{Sweep Dustpan} & \rotatebox{70}{Tray Oven} & \rotatebox{70}{Item Drawer} & \textbf{Avg.} \\
\midrule
RoboPrompt-SA & \twintabnum{94.0} & \twintabbestnum{100.0} & \twintabbestnum{82.0} & \twintabnum{83.7} & \twintabnum{64.7} & \twintabnum{78.7} & \twintabnum{58.7} & \twintabnum{18.0} & \twintabnum{44.0} & \twintabnum{11.7} & \twintabnum{89.3} & \twintabnum{26.3} & \twintabnum{37.3} & \twintabnum{60.6} \\
RoboPrompt-DA & \twintabnum{83.0} & \twintabbestnum{100.0} & \twintabnum{80.0} & \twintabnum{82.7} & \twintabnum{63.3} & \twintabnum{69.3} & \twintabnum{79.3} & \twintabnum{24.3} & \twintabnum{61.0} & \twintabnum{23.3} & \twintabnum{93.7} & \twintabnum{30.0} & \twintabbestnum{47.0} & \twintabnum{64.4} \\
Sequential Arms & \twintabnum{94.0} & \twintabbestnum{100.0} & \twintabnum{78.0} & \twintabnum{68.0} & \twintabnum{59.0} & \twintabnum{72.0} & \twintabnum{82.0} & \twintabnum{27.0} & \twintabnum{65.0} & \twintabnum{30.0} & \twintabnum{97.0} & \twintabnum{34.0} & \twintabnum{39.0} & \twintabnum{65.0} \\
\ours{} & \twintabbestnum{99.0} & \twintabbestnum{100.0} & \twintabnum{80.3} & \twintabbestnum{94.3} & \twintabbestnum{68.0} & \twintabbestnum{83.7} & \twintabbestnum{83.0} & \twintabbestnum{29.0} & \twintabbestnum{65.3} & \twintabbestnum{34.3} & \twintabbestnum{97.3} & \twintabbestnum{36.0} & \twintabnum{46.7} & \twintabbestnum{70.5} \\
\bottomrule
\end{tabular}%
}
\end{table}

\Cref{tab:ablation_sequential} shows that dimensionality reduction alone is beneficial but insufficient. Sequential Arms reaches $65.0\%$, improving over RoboPrompt-SA ($60.6\%$) and only marginally over RoboPrompt-DA ($64.4\%$), but remains $5.5$ percentage points below \ours{} ($70.5\%$). The gap is largest on tasks with strong role timing: Handover drops from $94.3\%$ to $68.0\%$, Handover Easy from $68.0\%$ to $59.0\%$, and Lift Ball from $83.7\%$ to $72.0\%$. This indicates that \ours{} is not merely easier because it halves the action dimensionality; the explicit trajectory-conditioning channel is needed to preserve inter-arm coordination.

\subsection{Test-Time Scaling Techniques}
\label{sec:ablation_conversation}

We evaluate two test-time scaling strategies commonly used with LLM agents. The first is \emph{Best-of-N}: we exploit generation stochasticity by executing the \ours{} pipeline $n{=}5$ times independently, producing candidate bimanual plans $\{\hat{\mathbf{A}}^{\text{bi}}_j\}_{j=1}^5$. Each candidate is then scored by a separate LLM-as-Judge call that compares the trajectory pattern against the in-context demonstrations and outputs a consistency score $s_j \in \{1,\ldots,5\}$; the final plan is selected as $j^*=\arg\max_j s_j$. This is related to self-consistency~\cite{wang2023selfconsistency}, adapted from majority voting to trajectory evaluation.

The second is \emph{Multi-Turn Conversation}. It starts from the same two calls as \ours{}: the leader predicts a single-arm trajectory, and the follower predicts its trajectory from the augmented prompt containing the leader plan. The variant then adds a refinement iteration. The previous turn is kept in the chat history, and we ask first the leader to revise its trajectory prediction based on the previous conversation. Second, also the follower is asked to adjust its trajectory based on the first turn and on the leader's revised plan. Thus, Best-of-N scales by sampling and judging independent plans, while Conversation scales by iterative cross-arm revision. Base \ours{} performs a single leader-follower pass without either mechanism.

\begin{table}[h]
\caption{\textbf{Test-time scaling: Conversation and Best-of-N.} Success rates (\%) on the TWIN benchmark.}
\label{tab:ablation_conversation}
\centering
\resizebox{\textwidth}{!}{%
\begin{tabular}{l|rrrrrrrrrrrrr|r}
\toprule
\textbf{Variant} & \rotatebox{70}{Push Box} & \rotatebox{70}{Dual Buttons} & \rotatebox{70}{Bottle Fridge} & \rotatebox{70}{Handover} & \rotatebox{70}{Handover Easy} & \rotatebox{70}{Lift Ball} & \rotatebox{70}{Lift Tray} & \rotatebox{70}{Pick Laptop} & \rotatebox{70}{Pick Plate} & \rotatebox{70}{Straighten Rope} & \rotatebox{70}{Sweep Dustpan} & \rotatebox{70}{Tray Oven} & \rotatebox{70}{Item Drawer} & \textbf{Avg.} \\
\midrule
\ours{} & \twintabbestnum{99.0} & \twintabbestnum{100.0} & \twintabnum{80.3} & \twintabbestnum{94.3} & \twintabnum{68.0} & \twintabnum{83.7} & \twintabnum{83.0} & \twintabbestnum{29.0} & \twintabnum{65.3} & \twintabnum{34.3} & \twintabnum{97.3} & \twintabbestnum{36.0} & \twintabbestnum{46.7} & \twintabnum{70.5} \\
+ Conversation & \twintabnum{93.0} & \twintabbestnum{100.0} & \twintabnum{74.0} & \twintabnum{15.0} & \twintabnum{21.0} & \twintabnum{78.0} & \twintabbestnum{92.0} & \twintabnum{27.0} & \twintabnum{47.0} & \twintabbestnum{36.0} & \twintabbestnum{99.0} & \twintabnum{33.0} & \twintabnum{32.0} & \twintabnum{57.5} \\
+ Best-of-N & \twintabnum{98.0} & \twintabbestnum{100.0} & \twintabbestnum{82.0} & \twintabnum{94.0} & \twintabbestnum{73.0} & \twintabbestnum{85.0} & \twintabnum{85.0} & \twintabbestnum{29.0} & \twintabbestnum{72.0} & \twintabnum{32.0} & \twintabnum{98.0} & \twintabnum{34.0} & \twintabnum{43.0} & \twintabbestnum{71.2} \\
\bottomrule
\end{tabular}%
}
\end{table}

\Cref{tab:ablation_conversation} shows that these two strategies behave very differently. Best-of-N slightly improves the average success rate from $70.5\%$ to $71.2\%$, with gains on Bottle Fridge ($80.3\% \to 82.0\%$), Handover Easy ($68.0\% \to 73.0\%$), Lift Ball ($83.7\% \to 85.0\%$), Lift Tray ($83.0\% \to 85.0\%$), Pick Plate ($65.3\% \to 72.0\%$), and Sweep Dustpan ($97.3\% \to 98.0\%$). However, the gain is small relative to its cost: Best-of-N requires $n$ leader-follower executions plus $n$ judge calls, and our latency analysis shows that it increases the token budget from ${\sim}21.5$k to ${\sim}180.7$k tokens per episode. Conversation, by contrast, degrades performance substantially. It reaches only $57.5\%$ average success, which is $13.0$ percentage points below \ours{} and $13.7$ percentage points below Best-of-N. The drop is concentrated on tasks that require precise role timing: Handover falls from $94.3\%$ to $15.0\%$, Handover Easy from $68.0\%$ to $21.0\%$, and Item Drawer from $46.7\%$ to $32.0\%$. Conversation improves only on Lift Tray ($92.0\%$), Straighten Rope ($36.0\%$), and Sweep Dustpan ($99.0\%$), where the gains are not sufficient to offset the failures on asymmetric coordination tasks.

To understand \emph{why} the conversational variant fails, we inspect the raw LLM predictions. The added refinement turns are intended to let each arm correct its trajectory after seeing the partner plan. Instead, they \emph{systematically corrupt the leader's predictions} in three ways:

\begin{enumerate}[nosep,leftmargin=*]
    \item \textbf{Gripper-state inversion.} In the handover tasks, the initial leader correctly reproduces the two-phase pattern from the demonstrations (gripper open during approach, then closed for transfer). After refinement, the first actions are flipped from open to closed and then reopened, inverting the grasp timing.
    \item \textbf{Spatial coordinate drift.} The leader's initial trajectory exhibits a clear y-coordinate phase transition (e.g., $y{=}30$ during approach, $y{=}45$ during transfer). In the refined plan, this transition is lost: all actions collapse to a single y-value, eliminating the spatial phase structure that the ICL examples encode.
    \item \textbf{Cross-arm coordinate leakage.} In severe cases, the x-coordinate of the refined leader shifts toward values characteristic of the \emph{follower's} workspace (e.g., $x{=}52 \to 30$), suggesting the model conflates the two arms' coordinate frames when attending to the accumulated conversational context.
\end{enumerate}

\noindent The follower's refined predictions, by contrast, remain largely unchanged from their initial values. The corruption is asymmetric and affects primarily the leader, whose initial plan was produced without seeing the partner's trajectory and is therefore most vulnerable to context-induced drift. These findings indicate that simply accumulating prior predictions in the chat history is harmful for this action-prediction interface: the extra context weakens the demonstration pattern and encourages cross-arm leakage rather than improving coordination.

\subsection{Including Rotations in Observations}
\label{sec:ablation_rotations}

As described in \Cref{sec:formulation}, our default observation representation consists of discretized 3D object centroids only. We evaluate an alternative that additionally includes Euler-angle rotations for each object, moving the per-object observation from $\mathbb{Z}^{3}$ to $\mathbb{Z}^{6}$.

\begin{table}[h]
\caption{\textbf{Ablation: Including rotations in observations.} \emph{Positions only} (default) includes only discretized 3D object centroids in the observation. \emph{+ Rotations} additionally includes the discretized Euler-angle orientation of each object.}
\label{tab:ablation_rotations}
\centering
\resizebox{\textwidth}{!}{%
\begin{tabular}{l|rrrrrrrrrrrrr|r}
\toprule
\textbf{Observations} & \rotatebox{70}{Push Box} & \rotatebox{70}{Dual Buttons} & \rotatebox{70}{Bottle Fridge} & \rotatebox{70}{Handover} & \rotatebox{70}{Handover Easy} & \rotatebox{70}{Lift Ball} & \rotatebox{70}{Lift Tray} & \rotatebox{70}{Pick Laptop} & \rotatebox{70}{Pick Plate} & \rotatebox{70}{Straighten Rope} & \rotatebox{70}{Sweep Dustpan} & \rotatebox{70}{Tray Oven} & \rotatebox{70}{Item Drawer} & \textbf{Avg.} \\
\midrule
Positions only (default) & \twintabbestnum{99.0} & \twintabbestnum{100.0} & \twintabbestnum{80.3} & \twintabbestnum{94.3} & \twintabnum{68.0} & \twintabbestnum{83.7} & \twintabbestnum{83.0} & \twintabnum{29.0} & \twintabnum{65.3} & \twintabbestnum{34.3} & \twintabnum{97.3} & \twintabbestnum{36.0} & \twintabbestnum{46.7} & \twintabbestnum{70.5} \\
+ Rotations & \twintabbestnum{99.0} & \twintabbestnum{100.0} & \twintabnum{78.0} & \twintabnum{77.0} & \twintabbestnum{69.0} & \twintabnum{73.0} & \twintabnum{76.0} & \twintabbestnum{32.0} & \twintabbestnum{72.0} & \twintabnum{19.0} & \twintabbestnum{98.0} & \twintabnum{22.0} & \twintabnum{32.0} & \twintabnum{65.2} \\
\bottomrule
\end{tabular}%
}
\end{table}

\Cref{tab:ablation_rotations} shows that adding rotations reduces the average success rate from $70.5\%$ to $65.2\%$ ($-5.3$ pp). The degradation is severe on Handover ($94.3 \to 77.0$), Straighten Rope ($34.3 \to 19.0$), Item Drawer ($46.7 \to 32.0$), Tray Oven ($36.0 \to 22.0$), Lift Ball ($83.7 \to 73.0$), and Lift Tray ($83.0 \to 76.0$). Only Pick Plate ($65.3 \to 72.0$) and Pick Laptop ($29.0 \to 32.0$) improve.

Rotations are obtained by fitting an Open3D Oriented Bounding Box (OBB) to the merged multi-view point cloud of each object, converting the OBB rotation matrix to xyz Euler angles, and discretizing at $5^{\circ}$ resolution into integers in $[0, 71]$. Each object observation thus doubles its dimensionality. Also here, we identify three factors behind the degradation:

\begin{enumerate}[nosep,leftmargin=*]
    \item \textbf{High rotation variance across demonstrations.} Inspecting the ICL prompts reveals that the rotation triple of randomly-oriented objects varies dramatically across the in-context examples; for the handover items, we observe orientations such as $[70,44,6]$, $[13,38,50]$, $[1,39,27]$, and $[52,35,17]$ for the same object class. In contrast, positions change smoothly from example to example. These erratic rotation values break the pattern structure that in-context learning relies on.
    \item \textbf{Euler-angle discontinuities.} Discretized xyz Euler angles are a poor metric space for SO(3): they suffer from gimbal lock and wrap-around at $0/72$, so two physically close orientations can map to very different integer triples. While the corresponding action space also uses discretized Euler angles, the action sequence exhibits a smooth trajectory that the LLM can extrapolate; observation rotations, by contrast, are unordered across demonstrations and thus appear as random noise.
    \item \textbf{Dilution of the positional signal.} Doubling the observation from 3 to 6 tokens per object lengthens the context and reduces the relative salience of the spatial coordinates that actually drive the task. This dilution effect is especially harmful for tasks with many objects (handover has five items, each contributing three extra tokens).
\end{enumerate}

\noindent The exception is Pick Plate ($+6.7$ pp): the plate consistently lies flat on the table, yielding stable rotation values across demonstrations (typically $[36, 36, \cdot]$), and its in-plane yaw encodes the approach angle needed for a successful bimanual grasp. This confirms that rotations \emph{can} be informative when they are \emph{consistent} across examples; for most tasks, however, the OBB-estimated Euler angles introduce more noise than signal, making position-only observations the better default.

\subsection{Increasing Discretization Resolution}
\label{sec:ablation_resolution}

Several low-performing tasks require fine spatial alignment: Pick Laptop involves sliding under a thin object, Straighten Rope requires contact near deformable endpoints, Tray Oven requires reaching into a constrained opening, and Item Drawer requires coordinated placement into a narrow container. We therefore evaluate whether a finer action and observation discretization improves performance. Specifically, we regenerate the ICL demonstrations and test observations using a $200^3$ voxel grid and $2.5^\circ$ rotation bins, while keeping the same leader-follower prompting, number of demonstrations, and evaluation protocol.

\begin{table}[h]
\caption{\textbf{Ablation: Spatial and rotational discretization.} Success rates (\%) on the four lowest-performing precision-sensitive tasks. The default setting uses a $100^3$ voxel grid and $5^\circ$ rotation bins; the high-resolution setting uses a $200^3$ voxel grid and $2.5^\circ$ rotation bins.}
\label{tab:ablation_resolution}
\centering
\begin{tabular}{l|rrrr|r}
\toprule
\textbf{Setting} & Pick Laptop & Straighten Rope & Tray Oven & Item Drawer & \textbf{Avg.} \\
\midrule
Default & \twintabnum{29.0} & \twintabnum{34.3} & \twintabbestnum{36.0} & \twintabnum{46.7} & \twintabnum{36.5} \\
High resolution & \twintabbestnum{34.0} & \twintabbestnum{35.0} & \twintabbestnum{36.0} & \twintabbestnum{52.0} & \twintabbestnum{39.3} \\
\bottomrule
\end{tabular}
\end{table}

\Cref{tab:ablation_resolution} shows a modest improvement from $36.5\%$ to $39.3\%$ average success over these four tasks. The gain is concentrated on Pick Laptop ($+5.0\%$) and Item Drawer ($+5.3\%$), where a finer positional grid can better represent thin grasp margins and drawer insertion targets. Straighten Rope changes only marginally ($+0.7\%$), and Tray Oven is unchanged, suggesting that their failures are not dominated by quantization alone. In these tasks, contact dynamics, occlusions, long-horizon replanning, and the LLM's ability to extrapolate consistent multi-step trajectories remain limiting factors.

Overall, higher discretization resolution partially alleviates precision bottlenecks but does not remove them. It also expands the integer range that the LLM must model from $[0,99]$ to $[0,199]$ for positions and from $[0,71]$ to $[0,143]$ for rotations, which can weaken the regularity of the token patterns in the ICL prompt. We therefore keep the $100^3$ voxel grid and $5^\circ$ bins as the default, and view finer discretization as a targeted option for tasks where quantization error is clearly the dominant failure mode.

\subsection{Point-Cloud Extraction Methods}
\label{sec:ablation_pointcloud}

Object positions are extracted from segmented point clouds across six RGB-D cameras and computing centroids (\Cref{sec:formulation}). For each object, the segmentation mask selects the relevant 3D points from each camera's depth-reconstructed point cloud. We compare three strategies for combining these per-camera point sets into a single centroid estimate:
\begin{itemize}[nosep,leftmargin=*]
    \item \textbf{Prune}: same as Concatenation, but applies a voxel downsampling step~\cite{open3d} (voxel size $0.02\,$m) before computing the centroid. Downsampling regularizes the point density across views, preventing cameras with denser depth maps from dominating the centroid estimate;
    \item \textbf{Standard}: computes the centroid of the segmented points independently in each camera view, then averages the per-camera centroids. This treats each viewpoint equally but is sensitive to cameras that see only a small or skewed portion of the object;
    \item \textbf{Concatenation}: concatenates all segmented points from all cameras into a single point cloud and computes the centroid over the merged set. This weights each view proportionally to the number of visible surface points, giving more influence to closer or less occluded viewpoints.
\end{itemize}

\begin{table}[h]
\caption{\textbf{Ablation: Point-cloud extraction method.} Average per-object Euclidean distance to ground-truth centroid (in cm, $\downarrow$ is better) for the three point-cloud extraction variants:}
\label{tab:ablation_pointcloud}
\centering
\resizebox{\textwidth}{!}{%
\begin{tabular}{l|rrrrrrrrrrrrr|r}
\toprule
\textbf{Method} & \rotatebox{70}{Push Box} & \rotatebox{70}{Dual Buttons} & \rotatebox{70}{Bottle Fridge} & \rotatebox{70}{Handover} & \rotatebox{70}{Handover Easy} & \rotatebox{70}{Lift Ball} & \rotatebox{70}{Lift Tray} & \rotatebox{70}{Pick Laptop} & \rotatebox{70}{Pick Plate} & \rotatebox{70}{Straighten Rope} & \rotatebox{70}{Sweep Dustpan} & \rotatebox{70}{Tray Oven} & \rotatebox{70}{Item Drawer} & \textbf{Avg.} \\
\midrule
Prune (default) & \twintabbestnum{4.56} & \twintabbestnum{1.10} & \twintabbestnum{2.36} & \twintabbestnum{0.87} & \twintabbestnum{0.92} & \twintabbestnum{6.08} & \twintabbestnum{0.28} & \twintabbestnum{0.19} & \twintabbestnum{1.09} & \twintabbestnum{0.49} & \twintabbestnum{11.02} & \twintabbestnum{11.32} & \twintabbestnum{4.52} & \twintabbestnum{3.45} \\
Standard  & \twintabnum{6.77} & \twintabnum{1.16} & \twintabnum{8.95} & \twintabnum{1.69} & \twintabnum{2.69} & \twintabnum{10.05} & \twintabnum{1.40} & \twintabnum{1.50} & \twintabnum{1.19} & \twintabnum{0.70} & \twintabnum{14.12} & \twintabnum{16.43} & \twintabnum{9.12} & \twintabnum{5.83} \\
Concatenation & \twintabnum{6.20} & \twintabnum{1.13} & \twintabnum{5.12} & \twintabnum{1.69} & \twintabnum{2.76} & \twintabnum{10.06} & \twintabnum{0.81} & \twintabnum{0.82} & \twintabnum{1.11} & \twintabnum{0.52} & \twintabnum{12.07} & \twintabnum{19.85} & \twintabnum{9.23} & \twintabnum{5.49} \\
\bottomrule
\end{tabular}%
}
\end{table}

\Cref{tab:ablation_pointcloud} reports the average Euclidean distance (in cm) between the estimated and ground-truth centroids. \emph{Prune} consistently achieves the lowest error across all 13 tasks, reducing the overall average from 5.83 (Standard) and 5.49 (Concatenation) to 3.45\,cm, a 41\% relative improvement over Standard. The gains are largest on tasks with small or partially occluded objects: Handover Easy ($2.69 \to 0.92$), Pick Laptop ($1.50 \to 0.19$), and Bottle Fridge ($8.95 \to 2.36$). We adopt Prune as the default throughout all experiments in the main paper.

\section{Simulation Qualitative Analysis}
\label{sec:supp_qualitatives}

\begin{figure}[H]
  \centering
  \includegraphics[width=0.99\linewidth]{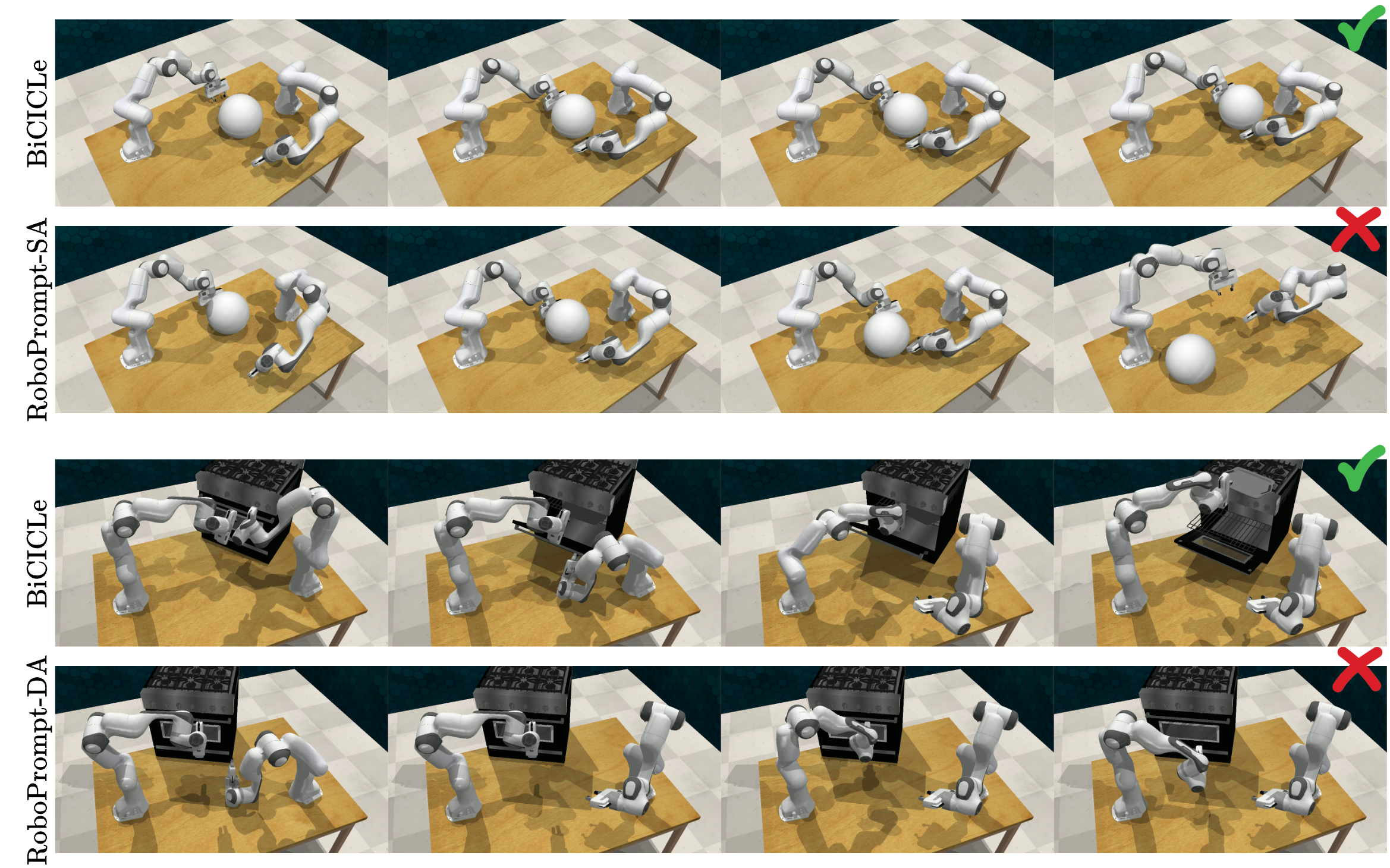}
  \caption{\textbf{Qualitative comparison.} Each pair of rows contrasts a successful \ours{} episode (\cmark) with a failed baseline episode (\xmark) on the same task. \emph{(Top two rows)} \textbf{Lift Ball} (tightly coupled symmetric): \ours{} vs.\ RoboPrompt-SA. \emph{(Bottom two rows)} \textbf{Tray Oven} (loosely coupled): \ours{} vs.\ RoboPrompt-DA. Columns show four keyframes sampled along the trajectory (left to right: initial approach, contact, manipulation, outcome).}
  \label{fig:qualitative}
\end{figure}

\Cref{fig:qualitative} contrasts \ours{} against the best-suited baseline architecture on two tasks from opposite ends of the coupling spectrum.

\myparagraph{Lift Ball (tightly coupled symmetric).}
In this task, both arms must approach the ball from opposite sides and simultaneously lift it. In the \ours{} episode (Row~1), the leader arm initiates contact from the left in Frame~2, and the follower mirrors the approach from the right, resulting in a balanced, symmetric grasp.
By Frame~4, both arms have risen in unison and the ball is stably lifted, as the explicit conditioning of the follower on the leader's trajectory enables the temporal synchronization necessary for this task.
In the RoboPrompt-SA episode (Row~2), the monolithic $\mathbb{Z}^{14}$ prediction struggles with per-arm precision: both arms converge toward the ball, but one arm applies contact with a slightly incorrect orientation. By Frame~4, the ball has rolled off the table. Despite SA's inherent advantage at capturing inter-arm correlation, the high dimensionality of the joint prediction space degrades the quality of individual arm trajectories. This is precisely the failure mode that the leader-follower decomposition avoids by halving the per-step prediction burden.

\myparagraph{Tray Oven (loosely coupled).}
This multi-step task requires one arm to open the oven door and the other to extract the tray, a loosely coupled interaction with a sequential dependency.
In the \ours{} episode (Row~3), the leader arm opens the oven door (Frames~1--2), and the follower, conditioned on the leader's completed action, reaches into the oven and grasps the tray (Frames~3--4). The sequential leader-follower structure naturally captures the temporal ordering of this task.
In the RoboPrompt-DA episode (Row~4), the two independent agents lack any information sharing. The arms move without coordination: the tray-grasping arm reaches into the oven while the door is not yet fully open (Frame~2), resulting in a collision and no possibility of extracting the tray. By Frame~4, the tray remains inside. This failure exemplifies how even loosely coupled tasks can require minimal temporal coordination that independent prediction fails to provide, despite DA's otherwise reasonable performance on this task category.

Overall, these qualitative examples confirm that our decomposition, a single, task-agnostic architectural choice, effectively bridges the dimensionality--coordination trade-off across the full spectrum of bimanual coupling, supporting the quantitative findings of \cref{sec:main_results} and establishing training-free ICL as a viable paradigm for multi-arm manipulation.

\section{RICL: Adapting a Vision-Language-Action Model for Bimanual ICL}
\label{sec:supp_ricl}

In the main paper, our evaluated LLM-based agents are \emph{training-free}: the LLM receives only text-based ICL demonstrations (visual-based only for the VLM-LF approach), and produces discretized actions without any gradient updates. A natural question is whether a \emph{trained} Vision-Language-Action (VLA) model can likewise benefit from in-context demonstrations in the bimanual setting. We investigate this through RICL (Retraining VLAs for In-Context Learning)~\cite{sridhar2025ricl}, which augments VLA inference with retrieved demonstrations.

\myparagraph{Architecture.}
RICL employs $\pi_0$-FAST-DROID~\cite{pertsch2025fast}, a flow-matching VLA based on $\pi_0$~\cite{BlackK-RSS-25} and fine-tuned on the DROID dataset~\cite{droid}, that predicts 15-step action chunks for a Franka Panda arm from three RGB camera views and the current joint state. Because $\pi_0$-FAST-DROID is a \emph{single-arm} policy, we instantiate two independent server processes: one retrieving ICL data for the right-arm agent, and one for the left-arm agent. To avoid drift at inference time, the agent queries both servers in parallel and integrates the predicted velocity chunks to obtain target joint configurations for each arm, which are executed by the simulator's built-in joint-position controller.

\myparagraph{In-context demonstrations for VLAs.}
Unlike text-based ICL, where textual demonstrations are serialized into the prompt, RICL prepends \emph{visual} demonstrations to the query observation. Each demonstration is a single timestep from a training episode and consists of three $224{\times}224$ RGB camera views (front, over-shoulder, wrist), the 8-dimensional proprioceptive state (7 joint positions + gripper), the corresponding action chunk (15 joint-velocity steps), and a language prompt. At inference, the front-camera view of the current observation is embedded with DINOv2 and matched against a FAISS index built over all training timesteps; the $N{=}4$ nearest neighbours are retrieved and prepended to the query observation before being fed to the VLA. The model is thus conditioned on visually similar demonstration contexts, analogous to the text-based ICL demonstrations used by our LLM-based agents.

\myparagraph{Bimanual adaptation.}
The dual-server setup mirrors the Dual Agent (DA) baseline from the main paper: the two arms predict independently with no explicit inter-arm coordination. The key difference is that RICL processes raw visual observations (three camera views per arm) rather than discretized text-based object positions, and outputs continuous joint-velocity trajectories rather than discretized keyframe actions.

\myparagraph{Relation to VLA policies.}
This comparison should not be interpreted as a claim that \ours{} is a drop-in replacement for end-to-end VLA policies. \ours{} is a structured keypose predictor: it assumes object-centric state estimates, predicts sparse discretized end-effector targets, and relies on a motion planner for execution. Standard VLA policies such as $\pi_0$ instead consume raw visual observations and produce continuous action chunks. For this reason, the most relevant reported $\pi_0$ baseline on TWIN is $\pi_0$-keypose from 3DFA~\cite{ze20253dfa}, which adapts the VLA to the benchmark's keypose prediction interface rather than evaluating the native action-output policy directly.

\myparagraph{Results.}
\cref{tab:ricl_results} reports per-task success rates for RICL compared with the training-free \ours{} base pipeline.

\definecolor{supgray}{gray}{0.88}
\begin{table}[h]
\caption{\textbf{VLAs vs.\ \ours{} on the TWIN benchmark.} \ours{} results from the main paper. \colorbox{supgray}{Gray:} supervised method (results from~\cite{ze20253dfa}).}
\label{tab:ricl_results}
\centering
\resizebox{\textwidth}{!}{%
\begin{tabular}{l|rrrrrrrrrrrrr|r}
\toprule
\textbf{Method} & \rotatebox{70}{Push Box} & \rotatebox{70}{Dual Buttons} & \rotatebox{70}{Bottle Fridge} & \rotatebox{70}{Handover} & \rotatebox{70}{Handover Easy} & \rotatebox{70}{Lift Ball} & \rotatebox{70}{Lift Tray} & \rotatebox{70}{Pick Laptop} & \rotatebox{70}{Pick Plate} & \rotatebox{70}{Straighten Rope} & \rotatebox{70}{Sweep Dustpan} & \rotatebox{70}{Tray Oven} & \rotatebox{70}{Item Drawer} & \textbf{Avg.} \\
\midrule
\rowcolor{supgray} $\pi_0$-keypose~\cite{ze20253dfa} & \twintabnum{93.0} & \twintabnum{38.0} & \twintabnum{22.0} & \twintabnum{2.0} & \twintabnum{59.0} & \twintabnum{97.0} & \twintabnum{72.0} & \twintabnum{27.0} & \twintabnum{41.0} & \twintabnum{7.0} & \twintabnum{2.0} & \twintabnum{68.0} & \twintabnum{40.0} & \twintabnum{43.7} \\
\midrule
\ours{} & \twintabbestnum{99.0} & \twintabbestnum{100.0} & \twintabbestnum{80.3} & \twintabbestnum{94.3} & \twintabbestnum{68.0} & \twintabbestnum{83.7} & \twintabbestnum{83.0} & \twintabbestnum{29.0} & \twintabbestnum{65.3} & \twintabbestnum{34.3} & \twintabbestnum{97.3} & \twintabbestnum{36.0} & \twintabbestnum{46.7} & \twintabbestnum{70.5} \\
RICL~\cite{sridhar2025ricl} & \twintabnum{59.0} & \twintabnum{6.0} & \twintabnum{9.0} & \twintabnum{2.0} & \twintabnum{14.0} & \twintabnum{73.0} & \twintabnum{32.0} & \twintabnum{2.0} & \twintabnum{4.0} & \twintabnum{6.0} & \twintabnum{15.0} & \twintabnum{11.0} & \twintabnum{30.0} & \twintabnum{20.2} \\
\bottomrule
\end{tabular}%
}
\end{table}

RICL achieves $20.2\%$ average success, substantially below both \ours{} ($70.5\%$) and $\pi_0$-keypose ($43.7\%$), the supervised keypose-specialized $\pi_0$ variant reported by 3DFA~\cite{ze20253dfa}. This gap is unsurprising given two compounding factors.

\emph{First, visual-based ICL is inherently limited.} VLM-LF, the visual-observation ICL baseline in the main paper, achieves only $13.4\%$, while RICL improves to $20.2\%$ by using a trained VLA backbone and retrieval over visual demonstrations. However, both remain far below the text-based \ours{} pipeline, indicating that conditioning on raw pixel similarity provides less task-relevant information than discretized text observations, particularly for bimanual tasks that require precise spatial coordination between arms.

\emph{Second, VLAs require benchmark-specific adaptation to perform well.} $\pi_0$-keypose uses the TWIN keypose interface, yet still falls short of the training-free \ours{} pipeline. RICL, which uses a DROID-trained checkpoint with no TWIN-specific training, faces a severe domain gap: the DROID dataset consists exclusively of single-arm Franka tabletop episodes, whereas TWIN features bimanual tasks with different object geometries, scene layouts, and dynamics. Despite $\pi_0$ being pre-trained on over 900 million timesteps of diverse robot data, the model cannot bridge this gap through retrieval-augmented ICL alone. This highlights a fundamental limitation of the VLA paradigm: when the deployment domain diverges from the training distribution, further adaptation is required, which directly undermines the appeal of in-context learning as a training-free adaptation mechanism.

\section{LLM Call Statistics and Inference Latency}
\label{sec:supp_latency}

In \Cref{tab:latency} we report the LLM call overhead for each agent variant on the Lift Ball task, using it as a representative example, with GPT-5-mini. Statistics are averaged over 100 evaluation episodes per variant. All measurements include network round-trip latency to the OpenAI API. For methods that issue multiple independent non-sequential calls, we implement those calls in parallel: this applies to all dual-agent baselines and to the candidate-generation and validation stages of Best-of-N.

\begin{table}[h]
\caption{\textbf{LLM call statistics per episode.} Calls, token counts, and wall-clock time for each agent variant on the Lift Ball task with GPT-5-mini. Wall-time reports the median with the interquartile range (IQR), the other columns report mean $\pm$ standard deviation. }
\label{tab:latency}
\centering
\resizebox{\textwidth}{!}{%
\begin{tabular}{l|c|c|c|c|c}
\toprule
\textbf{Variant} & \textbf{Calls/ep} & \textbf{Prompt tok/ep} & \textbf{Compl.\ tok/ep} & \textbf{Total tok/ep} & \textbf{Median wall-time/ep} \\
\midrule
KAT-SA & 5.0 $\pm$ 2.8 & 16\,141 $\pm$ 9\,168 & 14\,835 $\pm$ 8\,996 & 30\,976 $\pm$ 18\,000 & 242s (IQR 50.3--295s) \\
KAT-DA & 11.7 $\pm$ 4.6 & 28\,474 $\pm$ 11\,126 & 33\,718 $\pm$ 13\,838 & 62\,192 $\pm$ 24\,795 & 273.1s (IQR 238.1--307s) \\
RoboPrompt-SA & 1.6 $\pm$ 1.8 & 3\,025 $\pm$ 3\,403 & 4\,056 $\pm$ 3\,861 & 7\,081 $\pm$ 7\,194 & 41.5s (IQR 30.7--57s) \\
RoboPrompt-DA & 4.8 $\pm$ 5.3 & 5\,045 $\pm$ 5\,517 & 12\,463 $\pm$ 12\,519 & 17\,509 $\pm$ 17\,885 & 43.3s (IQR 35.2--79.5s) \\
\midrule
\ours{} & 3.8 $\pm$ 4.3 & 5\,908 $\pm$ 6\,599 & 15\,599 $\pm$ 17\,782 & 21\,507 $\pm$ 24\,334 & 125.3s (IQR 103.5--159.1s) \\
+ Conversation & 12.3 $\pm$ 11.3 & 21\,829 $\pm$ 20\,043 & 51\,998 $\pm$ 48\,401 & 73\,827 $\pm$ 68\,256 & 278.2s (IQR 246.8--587.5s) \\
+ Best-of-N & 33.5 $\pm$ 35.9 & 62\,421 $\pm$ 66\,629 & 118\,236 $\pm$ 127\,478 & 180\,657 $\pm$ 193\,967 & 149.7s (IQR 130.2--189.1s) \\
\bottomrule
\end{tabular}%
}
\end{table}

\myparagraph{Token efficiency.}
RoboPrompt-SA is the most token-efficient method at ${\sim}7.1$k total tokens per episode. Base \ours{} uses ${\sim}21.5$k tokens, which is comfortably below KAT-DA (${\sim}62.2$k). Conversation reaches ${\sim}73.8$k tokens, or about $3.4\times$ the base \ours{} budget, because each replanning step adds refinement calls with longer chat histories. Best-of-N reaches ${\sim}180.7$k tokens, or about $8.4\times$ the base \ours{} budget. The growth is driven by both more calls per episode ($3.8 \rightarrow 12.3 \rightarrow 33.5$ for \ours{}, Conversation, and Best-of-N) and by much larger completion-token counts, showing that the extra cost comes primarily from repeated trajectory generation and scoring rather than from prompt context alone.

\myparagraph{Wall-clock latency.}
RoboPrompt-SA is also the fastest method at ${\sim}41.5$s per episode, closely followed by RoboPrompt-DA at ${\sim}43.3$s. Base \ours{} requires ${\sim}125.3$s. Conversation has a much larger median wall-time of ${\sim}278.2$s and a wide IQR, because its refinement calls are sequentially dependent: later calls must wait for earlier arm predictions to be appended to the chat history. Despite being by far the most expensive in calls and tokens, Best-of-N has a median wall-time of ${\sim}149.7$s, much closer to base \ours{} than to Conversation or KAT-DA (${\sim}273.1$s). This gap is explained by parallelism.

\myparagraph{Cost--performance trade-off.}
On Lift Ball, Best-of-N achieves the best absolute performance at $85.0\%$, followed closely by base \ours{} at $83.7\%$. Conversation reaches $78.0\%$ on the same task, so its extra inference cost does not translate into better behavior. The key trade-off is therefore between the small accuracy gain of Best-of-N and its very large compute overhead: relative to base \ours{}, it improves success by only $1.3$ percentage points while increasing the token budget from ${\sim}21.5$k to ${\sim}180.7$k tokens per episode. Conversation is worse on both axes, increasing the token budget to ${\sim}73.8$k and median latency to ${\sim}278.2$s while reducing success. Relative to the RoboPrompt baselines, base \ours{} improves over RoboPrompt-SA ($78.7\%$) and RoboPrompt-DA ($69.3\%$), making the basic leader-follower pipeline the strongest efficiency--accuracy operating point, while Best-of-N is better viewed as a compute-heavy refinement for squeezing out the last few points of performance. A complementary way to reduce inference cost and latency is to use a smaller non-reasoning backbone, such as Qwen~2.5 7B: as shown in \Cref{tab:backbone}, the same architectural ranking is preserved, and \ours{} still surpasses the strongest training-free baselines on average.

\section{Prompt Templates}
\label{sec:supp_prompts}

\newcommand{\promptsep}{\texttt{\char`\>}}

We report the full prompt templates used by \ours{}.
Every agent call follows the standard \texttt{[system, user]} message format.
The \emph{system} message is fixed per agent, while the \emph{user} message is assembled at inference time by concatenating $N=10$ ICL demonstrations with the live observation (see format below).
Placeholders are typeset in \texttt{\textlangle angle brackets\textrangle}.

\subsection*{\ours{}}
\myparagraph{Step 1: Leader prediction.}

	\begin{tcolorbox}[breakable, colback=blue!3, colframe=blue!40!black, title=System prompt (leader arm), fonttitle=\bfseries\small, fontupper=\small]
	You are the \textlangle right$|$left\textrangle{} arm of a bimanual Franka Panda robot with parallel grippers.
	We provide you with some demos in the format of observation\promptsep{}[action\_1, action\_2, \dots].
	Then you will receive a new observation and you need to output a list of actions that matches the trend in the demos.
	Do not output anything else.
	\end{tcolorbox}

	\begin{tcolorbox}[breakable, colback=gray!5, colframe=gray!60!black, title=User prompt (leader arm), fonttitle=\bfseries\small, fontupper=\small]
	\textlangle demo$_1$ observation\textrangle{} \promptsep{} \textlangle demo$_1$ leader actions\textrangle, \dots, \textlangle demo$_N$ observation\textrangle{} \promptsep{} \textlangle demo$_N$ leader actions\textrangle, \textlangle live observation\textrangle{} \promptsep{}
	\end{tcolorbox}

\myparagraph{Step 2: Follower prediction.}

	\begin{tcolorbox}[breakable, colback=blue!3, colframe=blue!40!black, title=System prompt (follower arm), fonttitle=\bfseries\small, fontupper=\small]
	You are the \textlangle left$|$right\textrangle{} arm of a bimanual Franka Panda robot with parallel grippers.
	We provide you with some demos in the format of observation\promptsep{}[action\_1, action\_2, \dots].
	Then you will receive a new observation and you need to output a list of actions that matches the trend in the demos.
	Do not output anything else.
	\end{tcolorbox}

	\begin{tcolorbox}[breakable, colback=gray!5, colframe=gray!60!black, title=User prompt (follower arm), fonttitle=\bfseries\small, fontupper=\small]
	\textlangle demo$_1$ observation $\cup$ \{leader\_arm: demo$_1$ leader actions\}\textrangle{} \promptsep{} \textlangle demo$_1$ follower actions\textrangle, \dots, \textlangle live observation $\cup$ \{leader\_arm: predicted leader actions\}\textrangle{} \promptsep{}
	\end{tcolorbox}

\end{document}